\documentclass[a4paper]{article}

\usepackage[pages=all, color=black, position={current page.south}, placement=bottom, scale=1, opacity=0, vshift=5mm]{background}

\usepackage{caption}

\usepackage[margin=1in]{geometry} % full-width

\usepackage{amsmath}
\usepackage{amsthm}
\usepackage{amssymb}

\usepackage{float}

\newcommand{\TODO}[1]{\textcolor{blue}{#1}}

\newcommand{\OK}[1]{\textcolor{black}{#1}}

\usepackage[english]{babel}%,russian]{babel}

\usepackage[utf8]{inputenc}
\usepackage{hyperref}
\hypersetup{
	unicode,
	pdfauthor={Author One, Author Two, Author Three},
	pdftitle={A simple article template},
	pdfsubject={A simple article template},
	pdfkeywords={article, template, simple},
	pdfproducer={LaTeX},
	pdfcreator={pdflatex},
    pdfborder = {0 0 0} % УБИРАЕТ РАМКИ ВОКРУГ ССЫЛОК
}

\usepackage[sort&compress,numbers,square]{natbib}

\usepackage{graphicx, color}
\graphicspath{{fig/}}

\usepackage{algorithm, algpseudocode} % use algorithm and algorithmicx for typesetting algorithms
\usepackage{mathrsfs} % for \mathscr command

\usepackage{lipsum}

\title{Fast gradient-free activation maximization\\for neurons in spiking neural networks}
\author{
Nikita Pospelov$^1$\thanks{Corresponding author, \href{mailto:pospelov.na14@physics.msu.ru}{\texttt{pospelov.na14@physics.msu.ru}}\vspace*{2pt}}\quad\quad
\and Andrei Chertkov$^{2,3}$ \quad\quad
\and Maxim Beketov$^4$ 
\and Ivan Oseledets$^{2,3}$ 
\and Konstantin Anokhin$^1$}

\date{
    $^1$Laboratory of Neuronal Intelligence, Institute for Advanced Brain Studies,\\ 
    Lomonosov Moscow State University\\
    $^2$Artificial Intelligence Research Institute (AIRI)\\
    $^3$Skolkovo Institute of Science and Technology\\
    $^4$HSE University\\[1ex] Moscow, Russia\\[2ex]%
}

\begin{document}

% МАКСИМАЛЬНАЯ ШИРИНА ПОДПИСЕЙ К КАРТИНКАМ
\captionsetup{width=0.8\textwidth}

\maketitle

\begin{abstract}
        
\OK{Elements of neural networks, both biological and artificial, can be described by their selectivity for specific cognitive features. Understanding these features is important for understanding the inner workings of neural networks. For a living system, such as a neuron, whose response to a stimulus is unknown and not differentiable, the only way to reveal these features is through a feedback loop that exposes it to a large set of different stimuli. The properties of these stimuli should be varied iteratively in order to maximize the neuronal response. To utilize this feedback loop for a biological neural network, it is important to run it quickly and efficiently in order to reach the stimuli that maximizes certain neurons' activation with the least number of iterations possible.}

\OK{Here we present a framework with an efficient design for such a loop. We successfully tested it on an artificial spiking neural network (SNN), which is a model that simulates the asynchronous spiking activity of neurons in living brains. Our optimization method for activation maximization is based on the low-rank Tensor Train decomposition of the discrete activation function. The optimization space is the latent parameter space of images generated by SN-GAN or VQ-VAE generative models. To our knowledge, this is the first time that effective AM has been applied to SNNs.}

\OK{We track changes in the optimal stimuli for artificial neurons during training and show that highly selective neurons can form already in the early epochs of training and in the early layers of a convolutional spiking network. This formation of refined optimal stimuli is associated with an increase in classification accuracy. Some neurons, especially in the deeper layers, may gradually change the concepts they are selective for during learning, potentially explaining their importance for model performance.}

The source code of our framework, MANGO (for Maximization of neuronal Activation via Non-Gradient Optimization) is available on GitHub\footnote{\url{https://github.com/iabs-neuro/mango}}.\\ 
		

\end{abstract}

\tableofcontents

\section{Introduction}
\label{sec:intro}

\OK{Brain neurons respond selectively to certain properties of stimuli. For instance, neurons in the primary visual cortex are specialized on simple properties of visible images, such as the direction of motion \cite{HubelWiesel} or color \cite{BiPOLES}. Meanwhile, neurons from higher level visual areas specialize in more complex stimuli, such as the presence of a face within the visual field \cite{QuianQuiroga2023}. Identifying the specialization of living neurons involves varying the properties of a stimulus (e.g., the image presented to the neuron) iteratively to find those that elicit the most intense response \cite{ponce2019evolving, bardon2022face} - these approaches are known as \textbf{activation maximisation} (AM). However, to effectively solve this problem in vivo, one needs to:}
\OK{
\begin{itemize}
    \item Find an optimal stimulus in as few iterations and as accurately as possible.
    \item Investigate the stimulus space for multiple activation optima, including local and global optima.
\end{itemize} 
}

\OK{This is especially important considering the ethical implications of such studies in neuroscience \cite{neuroethics2019, neuroethics2023}. The faster and better one can obtain results, the less stress is imposed onto the studied animals, and fewer animals are needed. To address this challenge, it is necessary to develop and incorporate new mathematical optimization methods into experimental setups. This will allow to obtain the best approximations of the global activation optimum with a minimal number of iterations, meaning that the living object will be exposed to stimuli (such as images or sounds) for a shorter period of time.}

\OK{Nowadays, it is widely believed that the learning mechanisms of deep neural networks are significantly different from those in the brain \cite{Brain-is-not-like-ANN,Brain-is-not-like-ANN-2}. Therefore, all direct comparisons between the two should be made with caution.
However, modern artificial neural networks, despite their diverse architectures and training methods, exhibit similar selectivity towards complex stimuli \cite{ANN-neuron-specializations}. Interestingly, just like in living brains, ANN neurons can exhibit selectivity for multiple stimuli \cite{Multimodal-neurons-in-ANNs}.}

\OK{Since ANNs are essentially computational graphs of functions parameterized by the weights of layers, and modern computational hardware such as GPUs and TPUs are capable of efficiently computing both the value of the function at a point for inference and its derivatives with respect to parameters for backpropagation, it is relatively easy to ``dissect'' ANNs \cite{bau2017network}  and analyze the ``responsibility areas'' of each artificial neuron \cite{Olah-feature-visualization}.}

\OK{This makes ANNs ideal for applying activation maximization techniques based on gradients computation (see, for example, \cite{Nguyen2019} for a review). Gradient-free approaches are also applicable to this problem, such as those described in \cite{wang2022high}, where 15 different methods were compared for maximizing the response of a specific neuron in a deep ANN using a pre-trained generative adversarial network (GAN) \cite{GANpaper}. A comprehensive comparative analysis of various optimization techniques, including those based on genetic and evolutionary optimization, was conducted in \cite{TTOpt-paper}. An original approach, presented in the latter, based on the Tensor Train decomposition (TT decomposition, \cite{TT-Oseledets}) was found to have significant advantages in terms of speed, accuracy, and stability (given the same number of iterations).}

\OK{A more biologically realistic model of living neural networks is the family of artificial spiking neural networks (SNNs)\cite{SNNs-what-are-overview}, the activity of which unfolds in time – in contrast to conventional feedforward neural networks, which are mathematically just functions without any inherent temporal dynamics. SNNs represent an important step towards bridging the gap between the inner workings of artificial and biological neurons, as spike timing has been shown to play a crucial role in brain functioning \cite{Panzeri2001, AndradeTalavera2023}. With this biological relevance in mind, in present work we focus on the activation maximization of neurons in SNNs. Recently, the inner neuronal representations of data in SNNs have been studied and compared to those in ordinary artificial neural networks \cite{SNN-neural-representations-2023}.}

\OK{In our work we attempt to solve a related but different problem – finding effective stimuli (\textbf{MEIs for Most Exciting Images}, termed in \cite{Inception-what-excites-neurons-most})\TODO{,} specifically for SNN neurons. We aim to do this efficiently, with as few iterations as possible, and with a view towards future applications to biological neural networks. To the best of our knowledge, this is the first time that activation maximization algorithms have been applied to SNNs.
}

%changes in NN during learning, antropic interpretability paper, reservoir computing and last layers importance.

\OK{
This work aims to contribute to the developing field of interpretable AI and feature visualisation (\cite{erhan2009visualizing, DNN-visualization-Yosinski, samek2016-evaluating-visualizations-of-what-NN-learns, methods-for-interpreting-NNs}), as well as potentially provide useful tools for optimal stimulus search in biological experiments. By doing so, it aims to help establish connections between natural and artificial intelligence through a unified computational framework.
Our contributions in this paper can be summarized as follows:
\begin{itemize}
    \item We propose a new activation maximization method based on the low-rank Tensor Train decomposition. 
    \item We compare it to existing optimizers in the Nevergrad library \cite{Nevergrad} and show its better convergence to optimal stimuli and superior performance. We apply our method to learning the MEIs of a spiking ResNet convolutional neural network, which is the first time that AM methods have been applied to SNNs. 
    \item Furthermore, we studied the computed MEIs layer-wise throughout the network training and correlated model performance with the formation of highly selective neurons.
\end{itemize}
}

\subsection{Paper outline}
The paper is structured as follows.
Section \ref{sec:Background} provides the background:

\begin{itemize}
    \itemsep0em 
    \item \ref{subsec:AM} – on the overall activation maximization (AM) problem statement and context in living and artificial neural networks;
    \item \ref{subsec:SNNs} – on (artificial) spiking neural networks (SNNs), their brief history \& inner workings.
\end{itemize}
Section \ref{sec:Methods} introduces the methods of present work:
\begin{itemize}
    \itemsep0em 
    \item \ref{subsec:TT-decomposition} introduces the Tensor Train (TT) decomposition \& TT-based optimization methods;
    \item \ref{subsec:generative} is on the role of generative models in present work – mainly on Generative Adversarial Networks (GANs), for Variational Autoencoders (VAEs) see Appendix \ref{appendix:VAE};
    \item \ref{subsec:SNNs-we-used} is on certain SNN software implementation frameworks that we've tried and used;
    \item \ref{subsec:optimization-methods} is on optimization algorithms we propose to use and the ones we benchmark to;
    \item \ref{subsec:MANGO} is on the overall structure of our software framework, MANGO, and its prospects.
\end{itemize}
Section \ref{sec:results} provides the results \& findings of present work:
\begin{itemize}
    \itemsep0em 
    \item \ref{subsec:performance} demonstrates the performance of TT-based optimization methods on our problem;
    \item \ref{subsec:emergence-and-dynamics} contains the main results – on the emergence of neuronal specializations in layers of various depth and how these evolve with training time;
    \item \ref{subsec:complexity-and-diversity} demonstrates the results on the complexity \& diversity of neuronal specializations.
\end{itemize}
Section \ref{sec:Discussion} provides a discussion and broader context of present work:
\begin{itemize}
    \itemsep0em 
    \item \ref{subsec:neuronal-selectivity} discusses neuronal selectivity;
    \item \ref{subsec:limitations} outlines the limitations of present work;
    \item \ref{subsec:explainable-AI} discusses present work in broader context of explainable AI;
    \item \ref{subsec:in-vivo} provides an outlook on applicability of our framework for \textit{in vivo} AM.
\end{itemize}
Appendix \ref{appendix:MEI-gallery} presents notable examples of obtained Most Exciting Images (MEIs); appendix \ref{appendix:VAE}, as mentioned, gives more details on VAE-based generative models that we tried in our work.

\section{Background}\label{sec:Background}

\subsection{Activation Maximization in artificial and living neural networks}\label{subsec:AM}

\OK{Our goal in present work is to optimize the activation of a specific neuron in an artificial (ANN) or spiking (SNN) neural network by finding its most exciting input (MEI) stimulus. This problem lies at the intersection of contemporary neuroscience and deep learning (DL). In DL, well-known studies such as XDream \cite{XDream} and Inception Loops \cite{Inception-what-excites-neurons-most} have demonstrated how generative models can effectively map the latent space of stimuli (images) to perform activation maximization in ANNs. In neuroscience, recent research has shown that it is possible to study the function of neurons in living organisms by incorporating a generative model such as XDream into a feedback loop that exposes a living subject (a mammal with a well-developed visual system) to generated images. In \cite{ponce2019evolving}, XDream was used to identify the MEIs in the visual cortex of macaque monkeys. In \cite{bardon2022face} further exploration found that MEIs do not necessarily encode exact faces, but can also encode abstract objects (many of which look face-like).}

\OK{In present work, we aim to contribute to the rapidly developing field of integrating deep learning techniques into neuroscience to address the long-standing question of neural coding principles. We introduce a novel approach that efficiently solves the activation maximization problem, requiring fewer iterations for convergence, and does not require any information about the underlying system's structure.}

\OK{With the activation intensity/frequency of a certain neuron viewed, mathematically, as a function of the input ``fed'' to the NN, whether it is a biological or artificial neuron, the MEI is defined as}

\begin{equation}
    \textrm{MEI}_i\,\overset{\textrm{def}}{=} \underset{x\in S}{\arg \max}~A_i
    (x)
\end{equation}

\OK{where $S$ denotes the stimulus space, $A_i(x)$ is the activation function of $i$-th neuron in the system. In real-world applications, $A_i(x)$ is always measured with \textbf{noise} which complicates MEI determination. In practice, the MEI of a neuron is calculated iteratively. Additionally, due to constraints imposed by real-world experiments, we want to calculate $\textrm{MEI}_i$ as \textbf{efficiently} as possible using the least number of iterations possible. This is essential to save time and maintain the focus of the animal being studied in an \textit{in vivo} experiment, and to save computational resources in an \textit{in silico} experiment.}

\OK{When studying a biological system, such as an animal's brain, we don't have access to its internal structure and don't fully understand how information about stimuli is transmitted to specific neurons.
This is informally known as a \textbf{black box} function optimization problem - without explicit access to the gradient (derivatives) of the function.}

\OK{Optimization methods in general can be roughly divided into:}

\begin{enumerate}
    \item \OK{\textbf{Gradient-based} methods, dating back to Newton and his method, have evolved dramatically, with variations of Stochastic Gradient Descent (SGD) \cite{sgd} powering the training of millions of neural network-based machine learning models around the world today.}
    
    \item \OK{\textbf{Gradiend-free methods} that do not explicitly use information about the system's structure. There are several families of these methods, including deterministic and stochastic ones. Evolutionary algorithms, which have gained particular attention in the context of activation maximization in recent years \cite{wang2022high,Gradient-free-AM, XDream}, fall into this category.}
\end{enumerate}

Gradient-free methods are naturally more appropriate for the problem of AM in real-world systems due to activation being a black-box-like function. Of course, even if not explicitly accessible, the gradient can be approximated by finite differences, but 1) if the function's landscape is very ``wobbly'' – quicky changing – the error of gradient estimates will be effectively unbounded; and 2) noise that is present in measurements of function's values will also contribute to the error of gradient estimates.

In the context of AM, quite recently such gradient-free methods have been explored \cite{Gradient-free-AM}, and extensively compared to each other \cite{wang2022high} – both in silico and in vivo, with Covariance Matrix Adaptation (CMA) \cite{CMA} outperforming most in both settings.

In present work, we introduce (for the first time) another family of gradient-free optimization algorithms for the problem of SNN's neuron activation maximization – ones based on low-rank tensor decompositions (Tensor Train, \cite{TT-Oseledets}) of the optimized function. Some of these methods are essentially grid search for a maximum in an array, but a smart one, utilizing the low-rank tensor nature of the the array, thus achieving exponential speedup over straightforward search. Others belong to the Monte-Carly (MC) family of methods, which are less prone to being stuck in local optima by stochastically ``jumping'' around the array, not necessarily towards the optimum. See the corresponding section of Methods for a self-contained introduction of these methods.

\subsection{Spiking Neural Networks (SNNs)}\label{subsec:SNNs}

\OK{From a mathematical perspective, ordinary ANNs are simply parameterized functions that transform input data to produce output – such ANNs are called \textbf{feedforward}. Of course, the field of contemporary deep learning \cite{DL-Fathers-Nature-paper,DeepLearningBook} is rich in further complications of this idea, e.g. NN-based generative models can be said to be non-deterministic functions, but for the purposes of this discussion we limit ourselves to the basics. Unlike in ANNs, the activity of biological NNs \textbf{unfolds in time}. Living neurons are constantly accumulating signals from neighbour neurons they share synaptic connections with, and, if the total voltage amplitude of incoming signals (membrane potential, MP) exceeds some threshold, the neuron ``fires''. That is – it emits what's called a \textbf{spike}, a signal the form of which is very special (and well-studied \cite{Izhikevich-Spike-model}) due to the bio-chemical dynamics that governs its emission. The spike then travels along the synapse to neuron's neighbours – to provoke them to spike in their turn. This inherent temporal nature of biological NNs was overlooked when the first artificial models of living NNs, namely the 1943 neuron model of McCulloch \& Pitts \cite{Perceptron-McCulloch-Pitts} – further developed and implemented in hardware and termed Perceptron by Rosenblatt \cite{Rosenblatt-Perceptron-1,Rosenblatt-Perceptron-2} in 1957-58 – were introduced. Perhaps due to hardware limitations of the time, implementing feedforward functions lacking any oscillatory internal dynamics was more straightforward. The invention of error back-propagation algorithm \cite{Backpropagation-Hinton} in 1986 as a useful method for training such ANNs (e.g. to perform classification tasks after learning on exemplar data) further cemented the dominance of feedforward ANNs of feedforward nature as machine models of intelligence. It was Maass's work  \cite{Maass-SNN-origins} in 1997 that introduced networks of artificial spiking neurons as a model of biological NNs.}

\OK{An artificial spiking neural network (SNN), just like a biological one, can be thought of as a dynamical system, with each neuron being a subsystem with its own internal dynamics. Neurons are connected with each other, forming layers with connections being directional. Over time, a neuron is receiving signals from its sourcing neighbours. These signals are being summed with weights designating the importance of connections – to give the total value of ``membrane potential''. If this value exceeds a certain threshold (one of the hyperparameters of this model), the neuron ``fires'' – emits a spike into its output. For this to be effectively modelled digitally, this \textbf{dynamics is discretized}: the time-dependence of signal being encoded by an array of numbers (signal amplitudes). These arrays are basically a discrete model of what's called \textbf{spike trains} (term used for signals in biological NNs as well). This discretization is not too far-fetched as a model of biological reality: spikes in living neurons are almost ``binary'' – MP typically changes in almost discrete steps.}

% Temporal nature of spike trains in biological NNs is also nearly discrete – spikes have to all arrive at nearly the same time for their contributions to MP to add up.

\OK{Nowadays there are several software implementations of artificial SNNs, with at least two outstanding frameworks – \textbf{snnTorch} \cite{SNN-Torch-paper} \& \textbf{SpikingJelly} \cite{SpikingJelly-paper}, allowing to do complete training and inference procedures – these two were used in present work.}

\OK{There are several important questions to answer when working with an SNN. First, how is input data \textbf{encoded} for the SNN to process it. In our work we ``expose'' an SNN to static images. For that there are at least two approaches \cite{SNN-Torch-paper}: 
\begin{enumerate}
    \itemsep0em 
    \item simply repeating the image (unaltered) a certain number of frames (kind of a static video)
    \item showing the image on a certain number of frames, but altering it with some sort of noise.
\end{enumerate}
We've chosen the first approach for simplicity, so in our work the first spiking layer was \textbf{exposed to input image repeated for 20 to 100 frames}. Each frame is \textbf{first fed to a pre-trained (feedforward) convolutional layers} (for these to extract basic image features) with residual blocks. So the model we were working with is basically a spiking analogue of ResNet \cite{ResNet-paper}.}
%In that, the pre-trained CNN plays a role of the visual cortex (already formed and trained to extract visual primitives), while the SNN is somewhat like the associative cortex. For more details on our architecture, see further sections. 

\OK{The second basic question is: what exact \textbf{model of a spiking neuron} does one use? The above description is very schematic, one needs to specify a certain mechanistic model (an electric circuit, to be precise) of the living neuron, so that the dynamics of such a model accurately approximated reality. For practical applications, Hodgkin-Huxley model \cite{Hodgkin-Huxley-model}, while being most biologically (physically) accurate, is too complicated to implement and work with. So a practical choice is the one of so-called \textbf{Leaky Integrate-and-Fire (LIF) neuron models}, that date back to Louis Lapicque's 1907 work \cite{Leaky-Integrate-and-Fire-Lapicque}.}

\begin{figure}[H]
\centering
\includegraphics[width=0.8\textwidth]{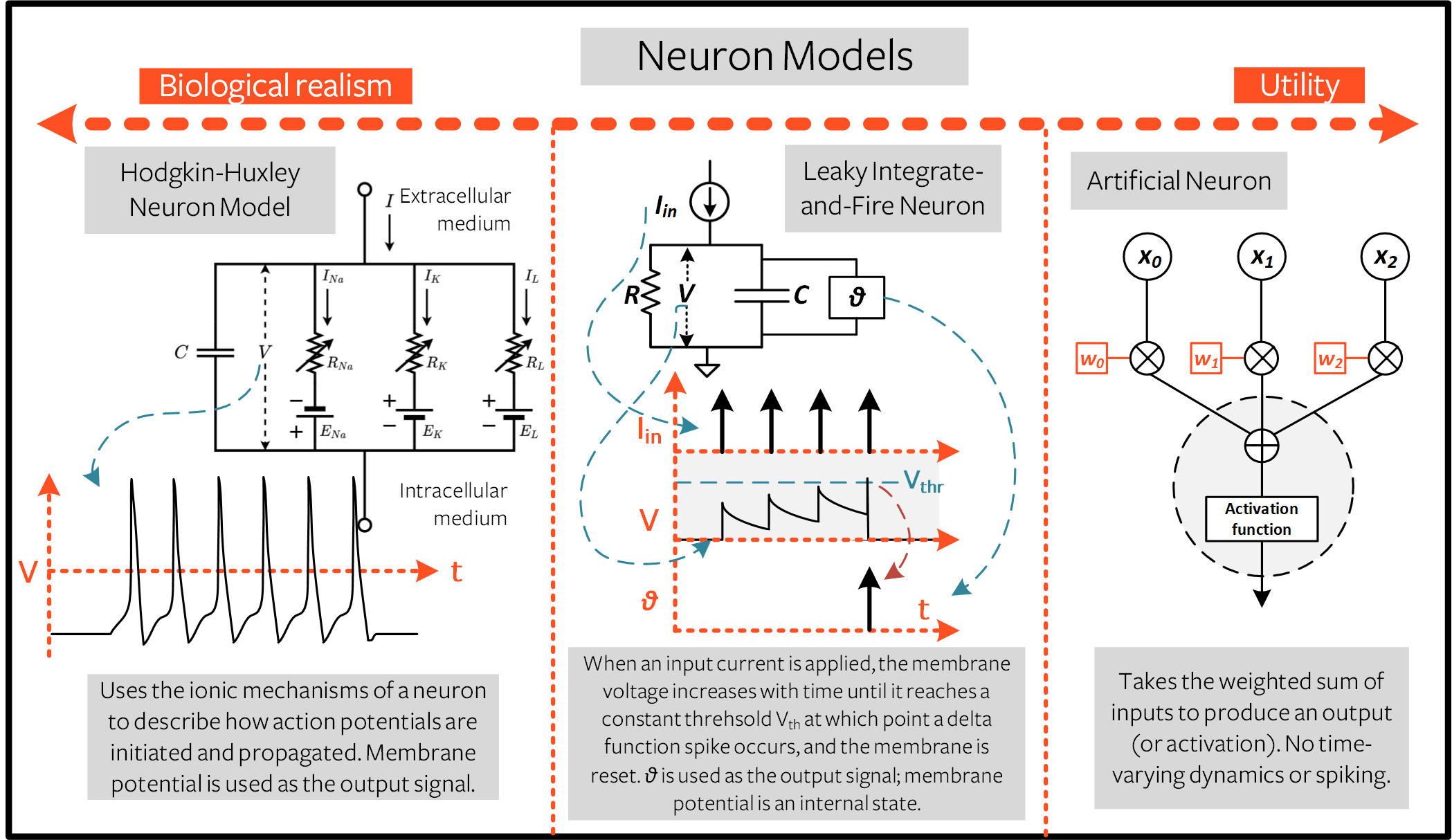}
\caption{Spiking neuron models (image from \href{https://snntorch.readthedocs.io/en/latest/tutorials/index.html}{snnTorch online tutorial} \cite{SNN-Torch-paper})}
\label{fig:spiking-neuron-models}
\end{figure}    

\OK{In LIF models, a spiking neuron is represented as an RC circuit – instead of directly summing the incoming (spike) voltages, such a neuron integrates input signal over time with leakage (due to R for resistance present in the circuit). A LIF neuron abstracts away the shape and profile of the output spike: it is simply treated as a discrete event. As a result, information is not stored within the spike, but rather the timing (or frequency) of spikes. Discrete-time equations of the spiking dynamics for such a neuron thus are:}

\begin{equation}
    \label{eq:spike-Heaviside}
    S[t] = \Theta(U[t]-U_\text{thr})=\begin{cases}
    1,\quad \text{if }U[t]>U_\text{thr}\\
    0,\quad\text{otherwise}
    \end{cases}
\end{equation}

\OK{with $S[t]$ being the signal (spike) intensity of a neuron at a discrete moment of time $t$, $U[t]$ being the MP. With $\Theta(x)$ being Heaviside's step function, a neuron discretely fires if its MP $U[t]$ exceeds a threshold value. The RC-circuit nature of a LIF neuron is described by the following (discretized) temporal dynamics equation on $U[t]$:}

\begin{equation}
    U[t\!+\!1] = \underset{\text{decay}}{\underbrace{\beta U[t]}}+
    \underset{\text{input}}{\underbrace{W X[t\!+\!1]}}-
    \underset{\text{reset}}{\underbrace{S[t] U_\text{thr}}}
\end{equation}

\OK{– at next moment of time, $t\!+\!1$, the MP is given by a linear combination (with learnable weights $W$) of inputs $X[t]$ from sourcing neigbhour neurons. There's also some decay (leakage) and a reset term. With $U_\text{thr}$ set to, say, 1, the only \textbf{hyperparameter} left to set is the decay rate, $\beta$.}

\OK{The third basic question is: how does one learn the weights of neurons' connections $W$? For that to be solvable with common machine learning approaches, the total loss function of SNN's activation on certain input data, $\mathcal{L}_W(\text{input})$, should be differentiable w.r.t. parameters of the network, $W$. The problem is that Heaviside's theta-function in Eq. \ref{eq:spike-Heaviside} is not differentiable at zero, and has zero derivative elsewhere. A traditional \cite{SNN-Torch-paper} way to overcome this is to smooth $\Theta$, replacing $S[t]$ with some \textbf{surrogate}, $\tilde{S}[t]$ – it could be any sigmoidal function – typically logistic function or arctangent. With $\tilde{S}$ being a smooth function, the NN's weights can be updated with gradient optimization methods. The choice of this surrogate function is another hyperparameter to be set when defining an SNN.}

\section{Methods}\label{sec:Methods}

\subsection{Tensor Train (TT) decomposition and TT-based optimization techniques}
\label{subsec:TT-decomposition}

\OK{As described above, our goal is to maximize the activation (spiking frequency) of certain neurons in an SNN trained to classify CIFAR-10 images. So we're optimizing a black-box function, the domain of which is the $\sim\!10^{2-3}$-dimensional latent space of image features (the latent representations of images are available with a generative model). If one picks a certain region of this domain and discretizes it (in our experiments, the latent space of images was \textbf{discretized into a 128-dimensional cube with 64 points on each side}), this high-dimensional tensor might be of quite low rank. The reasoning behind this is that since both ANNs (by design) and biological NNs (due to restrictions of the physical world) don't have exponential resources to encode all the places of interest in this space. Under such assumptions, optimization methods that are based on low-rank tensor decompositions might prove very effective.}

\OK{One kind of such decompositions that turned into a whole fruitful research field of its own is the Tensor Train (TT) decomposition, introduced in 2011 by Oseledets \cite{TT-Oseledets}. It allows to encode a low-rank high-dimensional tensor in a compact and convenient format, only using a polynomial (in dimension of the tensor) number of variables, instead of exponential required in general. The format resembles, if depicted graphically as what's called a tensor network, linked train cars, hence the name. In this format, tensors can be operated with: added, multiplied, convoluted, etc – very effectively (in the amount of operations and memory). All that provides a basis for applying TT-decomposition to solve various linear and nonlinear equations, PDEs, etc. Of many possible applications of TT-decomposition, other than to optimization problems, one could especially note ``tensorizing NNs'' \cite{Tensorizing-NNs,liu2022tt} – effectively building a compact tensor approximation to a nonlinear ANN. Apart from speeding up inference tasks, this seems very related to the subject of present work: the whole landscape of activation values of a certain neuron in an ANN could be compressed and investigated in detail. If the conjecture of low rank holds for activation functions of living neurons, this would mean that only a polynomial (in latent space dimension) number of ``requests'' (expositions to stimuli) is needed to effectively approximate the response of the living NN. This seems to be a promising direction of future research.}

\OK{There are thus several new optimization algorithms based on the TT-decomposition, where the (discretized) optimized function $f(x)$ is better and better approximated $\tilde{f}_{TT}(x)$ with a Tensor Train at each iteration of computing its values at different points. At the same time, the optimum candidate (optimum of $\tilde{f}_{TT}(x)$) is quickly found exploiting the structure of TT. This is the basic idea behind TTOpt optimization method \cite{TTOpt-paper}. Another related method is Optima-TT \cite{Optima-TT-Chertkov}. Also see \cite{Cichocki-tensors-for-optimization-1,Cichocki-tensors-for-optimization-2} for an overview of tensor approximation methods for optimization.}

\begin{figure}[H]
\centering
\includegraphics[width=0.8\textwidth]{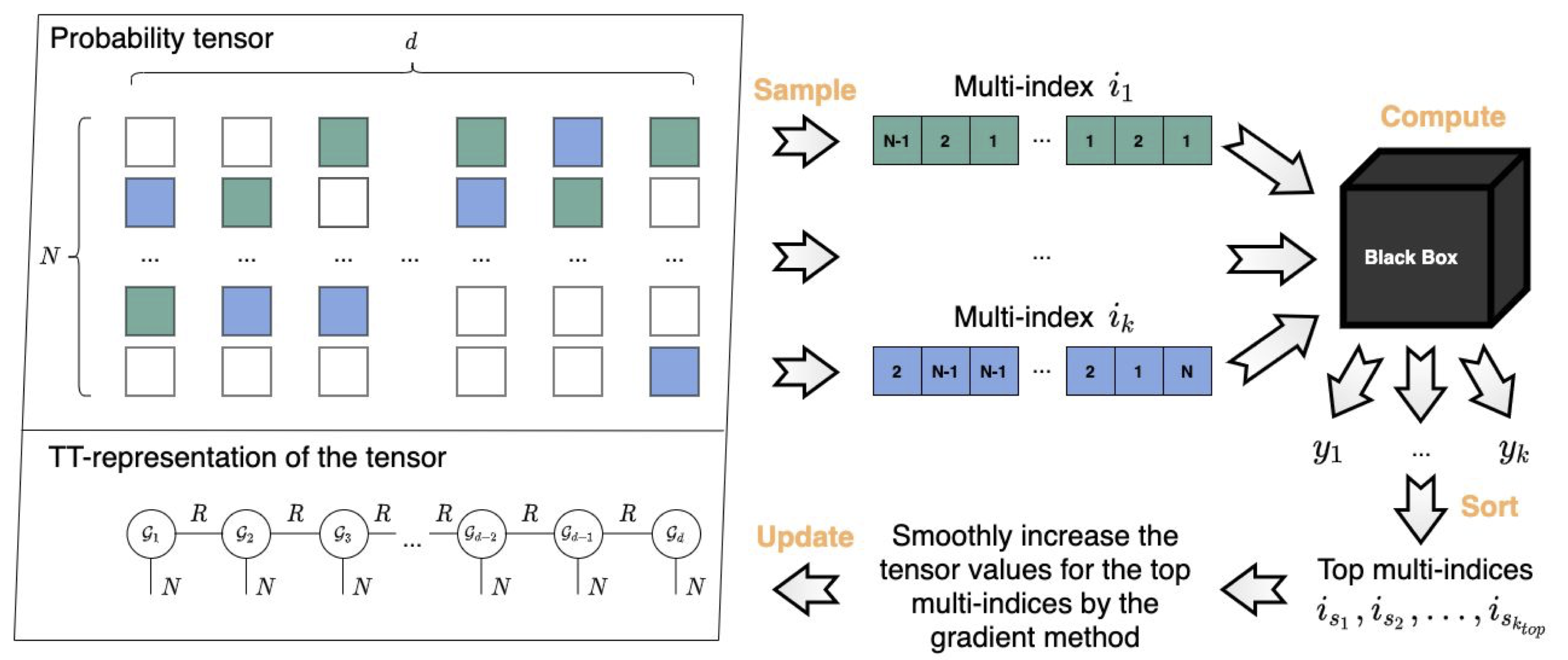}
\caption{Schematic of the PROTES method.}
\label{fig:PROTES-schematic}
\end{figure}

\OK{For our problem we've tried several TT-based optimization methods, and found PROTES \cite{PROTES-Chertkov-paper} to be the most effective. PROTES stands for ``Probabilistic Optimization with Tensor Sampling'' – it is a probabilistic optimization method. The main idea is similar to that of Simulated Annealing methods (see \cite{Simulated-annealing-what-is}) of the Monte-Carlo family. While gradient-based methods may be stuck in local optima, such methods have some probability of ``jumping'' out of these, if the temperature (yet another \textbf{hyperparameter}) of the ``optimum-seeking particles'' is high enough. The intuition behind PROTES is as follows: given a target function $f(\mathbf{x})$ to minimize (maximization done by minimizing $-\!f(\mathbf{x})$), apply the following monotonic transformation to it:} 

\begin{equation}\label{eq:fermi_dirac}
    F[f](x) = \frac1{
        \exp \bigl(
            (f(\mathbf{x}) - y_{\text{min}} - E) / T
        \bigr)
        + 1
    }.
\end{equation}

\OK{(In physics this expression is known as Fermi-Dirac statistic.) With $y_\text{min}$ being an exact or approximate minimum of $f$, $T>0$ - the ``temperature'' parameter, and $E$ – some ``energy'' threshold. With $F$ being a CDF-like function, one can stably find the maximum of its expectation: $\max_\theta \mathbb{E}_{\xi_\theta} F[f](\xi_\theta)$ where a family of random variables $\xi_\theta$ has a parametric distribution with density $p_\theta(\mathbf{x})$. Using the REINFORCE trick \cite{REINFORCE-trick}, one can estimate the expectation gradient as}

\begin{equation}
    \nabla_\theta\mathbb{E}_{\xi_\theta}F[f](\xi_\theta)\approx \frac1{M}\sum_{i=1}^M 
    F[f](\mathbf{x}_i)\nabla_\theta \log p_\theta(\mathbf{x}_i)
    \label{expectation-gradient}
\end{equation}

\OK{with Monte-Carlo – $\{\mathbf{x}_i\}_1^M$ being i.i.d. realizations of the r.v. $\xi_\theta$. If one manages to find optimal parameter values $\tilde{\theta}$ of $p_\theta$, then $p_{\tilde{\theta}}$ is expected to have a peak at the maximum of $F[f]$. At ``low temperature'' (small values of $T$), only a few terms contribute to the sum (\ref{expectation-gradient}) – namely, with those $\mathbf{x}_i$ for which $f(\mathbf{x}_i)-y_\text{min}<E$ (``low energy particles'') – for those, $F[f]\approx 1$, while for others it is close to zero. Thus one keeps only a few best values of the sample. With $p_\theta$ having a low-rank TT-representation, the above procedure (sampling and finding top-n values of the array) can be done quickly and effectively.}

\subsection{Generative modeling}\label{subsec:generative}

\OK{The task of a generative model in the pipeline of our experiment is to effectively ``map'' the space of stimuli, providing lower-dimensional latent coordinates for the subspace of ``natural'' images in the huge space of all possible images of given size. The models were trained on CIFAR-10  \cite{CIFAR10-dataset-reference} – a dataset of 60.000 color images of size 32x32 pixels, split into 10 classes (6.000 images per class): 0) airplane, 1) automobile, 2) bird, 3) cat, 4) deer, 5) dog, 6) frog, 7) horse, 8) ship, 9) truck – see Fig.\thinspace\ref{fig:CIFAR10-classes} in Appendix\thinspace\ref{appendix:VAE} for examples.}

\OK{We've tried two types of generative models that are known to perform well on this task:} 

\begin{enumerate}
    \itemsep0em 
    \item Generative Adversarial Networks (GANs) \cite{GANpaper}, specifically their spectral-normalized version (SN-GAN) \cite{SN-GAN-paper}

    \item Variational AutoEncoders (VAEs) \cite{VAE-vanilla-Kingma-Welling}, specifically their ``discretized'' (vector-quantized) version – VQ-VAEs \cite{VQ-VAE-Van-Den-Oord}
\end{enumerate}

\OK{We've found SN-GANs more suitable in our setting – see Appendix \ref{appendix:VAE} for a review on VQ-VAEs and comparison of both models performance on our task. That is why in this section we only provide an overview of GANs.}

\OK{A Generative Adversarial Network (GAN) \cite{GANpaper} is comprised of two NNs, a generator $G$ and a discriminator $D$, playing a minimax game: $G$ has to learn a distribution close to that of natural examples (images), fooling $D$ that they are realistic; $D$ has to discriminate images produced by $G$ from real ones – by learning the probability distribution of the latter (supported on the space of latent features of the images).} 

\OK{Due to the unstable nature of this minimax game $G$ and $D$ are playing, original GANs do not perform exceptionally well on data such as CIFAR-10, so quite soon after their introduction, many regularization techniques were proposed. Quite popular one being the Spectral Normalization (SN-GAN) \cite{SN-GAN-paper} – with the layers' weight matrices being penalized so that their Lipschitz constant is bounded from above. This simple yet elegant solution allowed the authors of SN-GAN reach then-state-of-the-art results on CIFAR-10 in terms of metrics of inception score \cite{GAN-metric-inception-score} and Fr\'echet inception distance (FID) \cite{GAN-metric-Frechet}. Subjectively, the quality of images generated by an SN-GAN trained on CIFAR-10, is very respectable.}

\OK{After running several experiments, we've settled on using an SN-GAN with the \textbf{latent space of dimension 128}, with each dimension discretized into 64 points.}

\subsection{SNN architectures}\label{subsec:SNNs-we-used}

\OK{Before moving on to experiments with maximizing the neuronal responses \textit{in vivo}, it is necessary to fully prepare and test the pipeline of such an experiment, where living neurons are replaced by some adequate \textit{in silico} model. The most studied class of such models are artificial spiking neural networks (SNN). Among all the software implementations of SNN, two most well-known and developed one were chosen - SNNTorch \cite{SNN-Torch-paper} and SpikingJelly \cite{SpikingJelly-paper} libraries.} 

\OK{What is special about spiking neural networks, both artificial and living ones – is that the spikes are discrete in time. So it is not immediately obvious how to backpropagate error through the output of an artificial SNN, as noted in section \ref{subsec:SNNs}. In present work, we use the solution introduced in \cite{SNN-Torch-paper}, based on the so-called surrogate gradient method, see Fig.\thinspace\ref{fig:spiking-neuron-models}.}

\OK{First, a simple convolutional NN based on SNNtorch was implemented and trained to solve the problem of image classification from the CIFAR-10 dataset (10 classes of color images of 32x32 pixels). Model architecture consisted of ``ordinary'' (feedforward, not spiking) convolutional layers, as well as fully-connected layers. The spiking nature of the model was achieved by projecting the input of ordinary layers to the \textbf{Leaky Integrate-and-Fire neurons with a membrane decay constant of 0.9}.}

\OK{The resulting SNN was trained to achieve \textbf{classification accuracy of 72\%} on the CIFAR-10 dataset. This number is lower than that of state-of-the-art feedforward (not spiking) architectures, but we considered it quite sufficient for our preliminary studies.}

\OK{Next, to speed up inference and increase network depth, we integrated SpikingJelly \cite{SpikingJelly-paper} into our framework. We used a spiking analogue of a celebrated ResNet18 model \cite{ResNet-paper} for the main results of present work, since this architecture provided an optimal trade-off between depth and inference speed. Again, we {used LIF neurons with membrane decay constant equal to 0.9}. The trained network achieved \textbf{86\% accuracy} on CIFAR10 dataset.}

\OK{Single neuron activity was determined by the number of spikes that it produced during ``exposure'' time (time was \textbf{discretized into 100 counts for our SNNtorch-based network and 50 for SpikingJelly-based one}). Activity was averaged over all feature channels and normalized to the range [0,1].}

\subsection{Optimization algorithms}\label{subsec:optimization-methods}

\OK{
One of the most popular and effective software packages for gradient-free optimizaton is the Nevergrad framework \cite{Nevergrad}. Methods implemented there were used as benchmarks for our research. One distinctive feature of  Nevergrad is that it implements techniques that combine several optimization approaches into one, called ``Portfolios''.}

\OK{Based on our testing, we found that the basic Portfolio method, which combines methods of 1) covariance matrix adaptation evolution strategy (CMA-ES), 2) differential evolution (DE), and 3) SCR-Hammersley, performed the best in our problem setting. We also compared it to other popular gradient-free methods implemented in Nevergrad, such as 1) OnePlusOne, 2) NoisyBandit, and 3) simultaneous perturbation stochastic approximation (SPSA), which all showed poorer results compared to the default Portfolio. It should be noted that in \cite{wang2022high}, it was demonstrated that the CMA-based optimization method outperformed the maximum activation of genetic algorithms (by 66\% in silico and 44\% in vivo), which are popular choices for non-gradient-based AM tasks.}

% здесь бы в будущем вставить таблицу со сравнением

\OK{As an alternative to Nevergrad algorithms, we used PROTES optimization method \cite{PROTES-Chertkov-paper} (described in previous sections), based on the low-rank Tensor Train decomposition \cite{TT-Oseledets}. In addition to baseline PROTES, we applied quantization of the tensor modes (transforming each tensor dimension into a new set of dimensions with smaller modes), which allowed us to achieve better results compared to the baseline PROTES method.}

To summarize, the optimization techniques tested were:
\begin{enumerate}
    \itemsep0em 
    \item (baseline) Portfolio from the Nevergrad package
    \item (baseline) PROTES with $K=10$; $k_{top}=1$
    \item TT-s = PROTES with $K=5$; $k_{top}=1$ + mode quantization
    \item TT-b = PROTES with $K=25$; $k_{top}=5$ + mode quantization
    %\item TT-exp = PROTES with $k=100$; $k_{top}=10$ + quantization
\end{enumerate}

\OK{For PROTES, $K$ stands for size of the sample of optimum candidates, sampled from the model probability density; and $k_{top}$ is the number best candidates out of $K$, based on which the parameters of this model density are updated on the last step of each iteration}.

\subsection{MANGO framework}\label{subsec:MANGO}

\OK{We have developed a software framework for fast and accurate calculation of MEIs in artificial neural networks - called MANGO (Maximization of neuronal Activation via Non-Gradient Optimization). We were particularly interested in considering models that are more closely related to the functioning mechanisms of biological neurons, specifically spiking neural networks (SNNs), as one of our main future goals is to apply these methods to living systems. However, the gradient-free optimization techniques we propose in this paper are also well suited for classical neural networks.}

\begin{figure}[H]
\centering
\includegraphics[width=0.7\textwidth]{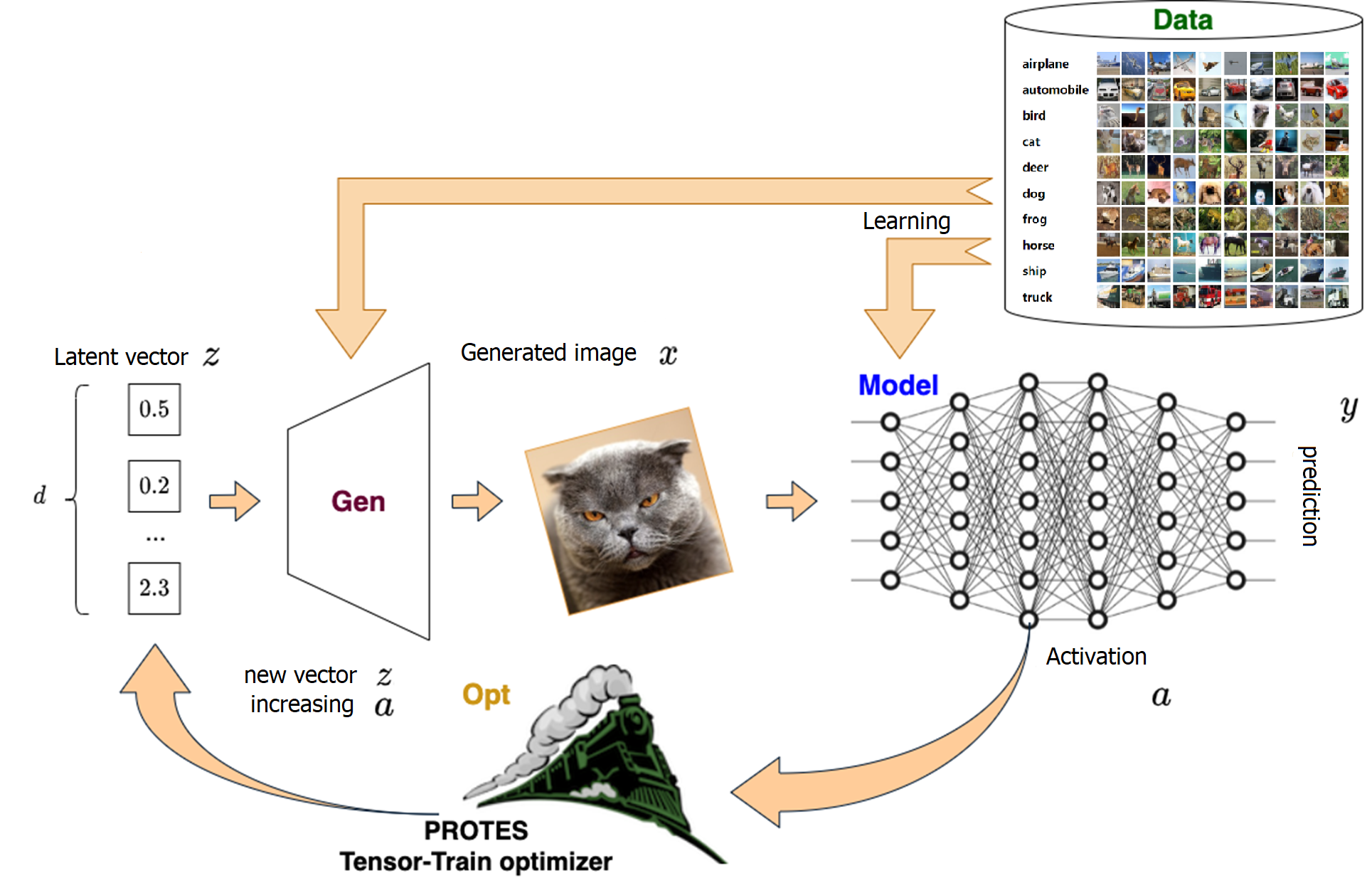}
\caption{Schematic of the MANGO framework.}
\label{fig:MANGO-framework}
\end{figure}

\OK{Within the framework, one can select a dataset, a generator model, a target neural network model, and an optimization method. One can also generate MEI search using various hardware backends and save and analyze the results.}

The following list shows the options that have been added to the framework:
\begin{enumerate}
    \itemsep0em 
    \item Datasets: MNIST \cite{MNIST}, Fashion-MNIST \cite{Fashion-MNIST}, CIFAR10 \cite{CIFAR10-dataset-reference}, Imagenet \cite{Imagenet}
    
    \item Generative models: VQ-VAE \cite{VQ-VAE-Van-Den-Oord}, SN-GAN \cite{SN-GAN-paper}

    \item Classic neural networks: AlexNet \cite{AlexNet}, Densenet \cite{Densenet}, VGG \cite{VGG=VeryDeepCNN} 
    
    \item Spiking neural networks: SNNTorch \cite{SNN-Torch-paper}-supported model convolutional networks and SpikingJelly-supported spiking ResNet18 \cite{SpikingJelly-paper, ResNet-paper}

    \item Optimization methods: Nevergrad-based benchmarks \cite{Nevergrad} and Tensor Train decomposition-based methods (TTOpt, PROTES, etc.) \cite{TTOpt-paper,PROTES-Chertkov-paper}

    \item backends: CPU (with multithreading), GPU, CuPy for SNN-based models
    
\end{enumerate} 
 
The selected dataset is used to train a convolutional (in our case, also spiking) network for the image classification task, as well as to train a generator network for the task of compressing input data into an effective latent representation. After training, the generator creates random latent representations, which are transformed into images and presented to the convolutional network. The activation of the neuron of interest is measured and fed as input to the optimizer, which produces an improved latent vector that maximizes the likely response of the neuron studied. This process is repeated multiple times until convergence or until the optimizer query budget is exhausted.

MANGO code is available on GitHub\ \url{https://github.com/iabs-neuro/mango}, including training and analysis scripts. Full sets of MEIs will be provided to interested readers upon request.

\section{Results}\label{sec:results}

\subsection{Performance of Tensor Train-based optimization methods}\label{subsec:performance}

\begin{figure}[H]
\centering
\includegraphics[width=0.8\textwidth]{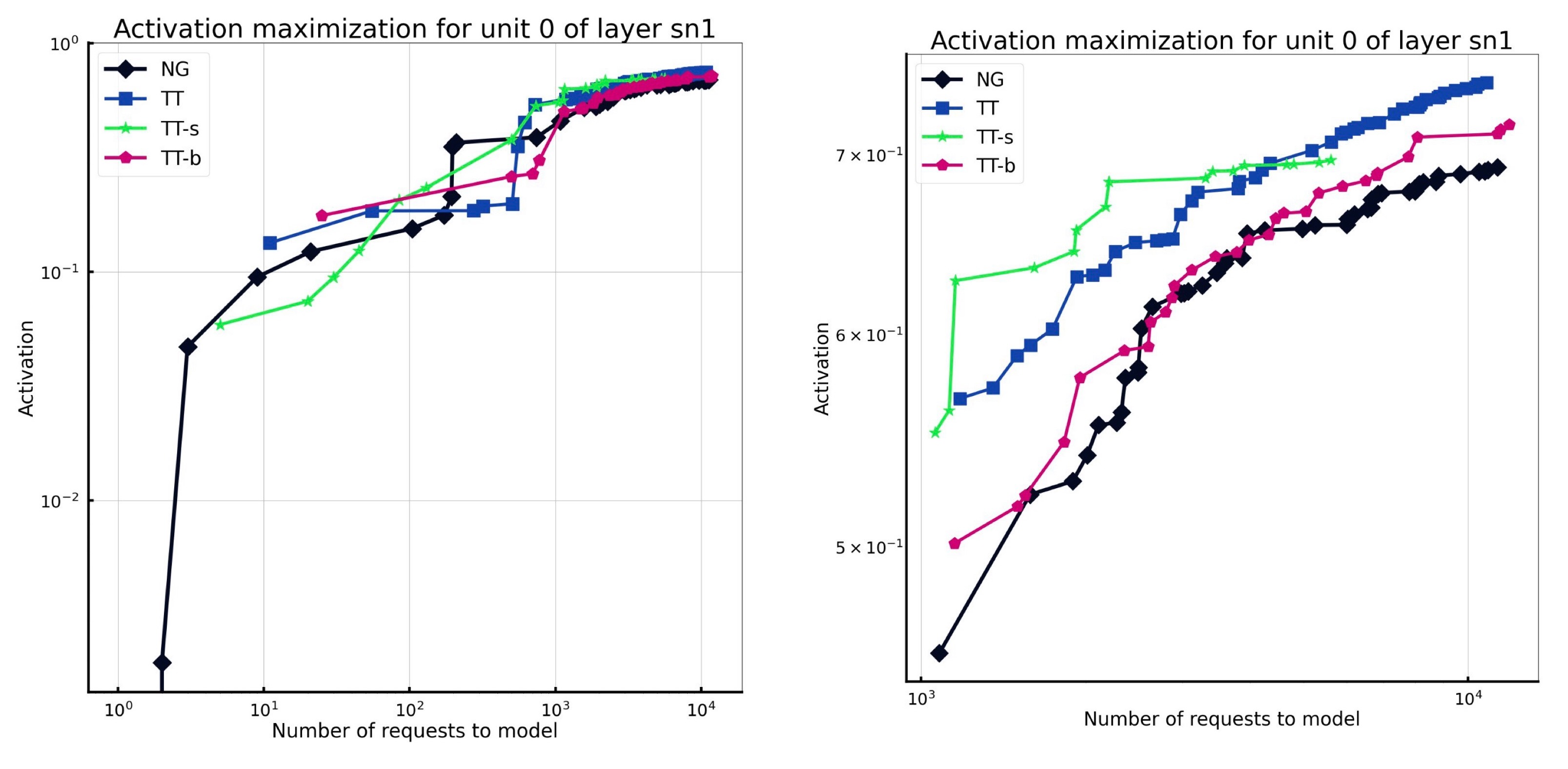}
\caption{{Activation of one selected neuron (unit 0, first LIF spiking layer, spiking ResNet18) depending on the number of requests to the optimizer. \textbf{Left}: full optimization history, \textbf{Right}: inset for optimization budget from 1000 to 12000.}}
\label{fig:opt-conv-4methods}
\end{figure}

\OK{Among the methods tested, approaches based on Tensor Train decomposition demonstrated performance as good or \textbf{10-20\% better} than the benchmarks (mainly the baseline Nevergrad Portfolio method, which was the most effective among the non-gradient methods considered, see Fig.\ref{fig:opt-conv-4methods}). Performance was measured in terms of target neuron activation in response to generated MEI.}

\OK{The PROTES method, in various modifications, has been shown to be 3-5 times faster than the Nevergrad Portfolio benchmark, depending on the parameters and properties of the image. At the same time, tensor methods require fewer steps to achieve a high level of activation of the target neuron (see Fig.\thinspace\ref{fig:opt-conv-4methods}).}

\OK{
Three different (although related) tensor methods and a non-gradient benchmark from the Nevergrad library often converged on the same images. This is surprising, given the vast number of image variations sampled from the generator's latent space. Despite the fact that the differences between MEIs increased in deeper layers (see Section \ref{subsec:complexity-and-diversity}), they were very similar in the early layers of the network (see Figs\thinspace\ref{fig:MEI-gallery-0}, \ref{fig:argmax-gen-comp-color}, \ref{fig:argmax-gen-comp-stripes}).}

\OK{
This suggests that all discrete optimization algorithms indeed converge to a good optimum in the latent space of the generator. Considering the results from Appendix \ref{appendix:VAE}, we can extend these findings to the general image space and conclude that discrete optimization techniques, especially those based on Tensor Train decomposition, are effective in finding MEIs.}

\subsection{Emergence and dynamics of neuronal specializations}\label{subsec:emergence-and-dynamics}

\begin{figure}[H]
\centering
\includegraphics[width=0.7\textwidth]{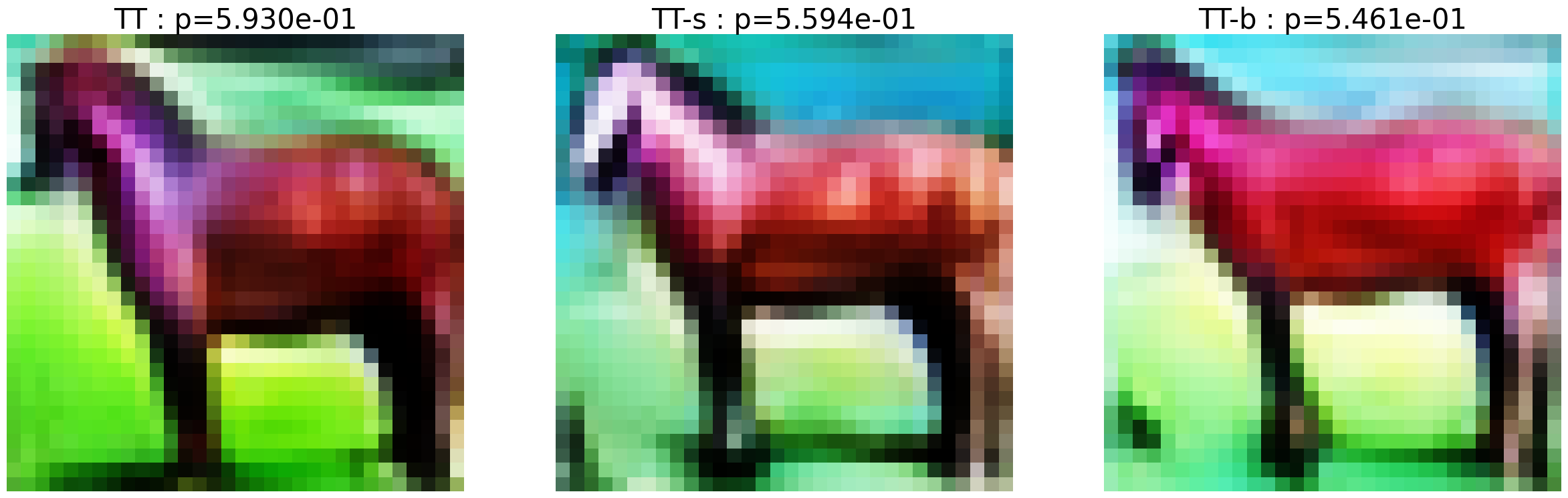}
\caption{``Pink horse'' neuron 52 from LIF layer 1.1 of spiking ResNet18. Images were generated using Tensor Train based methods. Numbers show activations of the target neuron on MEI}
\label{fig:argmax-pink-horse}
\end{figure}

\OK{To study the dynamics of MEI emergence and their content we analyzed optimal stimuli for the Resnet18 SNN spiking layers. We were interested in the maximal neuronal activations achieved during the optimization process and their correspondence with objects in the training images. The training was conducted for 1.000 epochs and the network states were recorded after each epoch. Nine spiking layers were chosen, evenly spaced between the input and output layers, with 64 neurons in each layer selected for MEI construction. Neuronal activation was maximized for all 576 considered neurons sequentially by searching for the most exciting visual input using three variants of the PROTES method based on TT decomposition. SN-GAN was used as the generative model (we also used VQ-VAE image generator, the comparison between these two models is given in Appendix \ref{appendix:VAE})}.

%in some experiments, different random initializations of the TT-exp method were used to save computational time \TODO{check}

\OK{An important aspect when analyzing MEIs is the relationship between identified neuronal specializations and the patterns the network has learned from the dataset.}
\OK{To study this, the resulting MEIs were then fed into the network to determine if the target neuron was specific to a particular image class. To quantify neuronal selectivity for specific classes, we've run the network on the MEIs and obtained the model class probabilities. Arguably, a more uniform class probability distribution corresponds to more abstract and general images. Conversely, a high probability for a specific class indicates the presence of that respective pattern in the MEI. For each MEI, we calculated the entropy of the final class probability distribution and normalized it against the maximum possible entropy (for a uniform distribution over 10 classes).}

%The procedure was repeated 100 times to account for single-trial spike variability inevitably present in SNN.

\begin{figure}[H]
\centering
\includegraphics[width=0.8\textwidth]{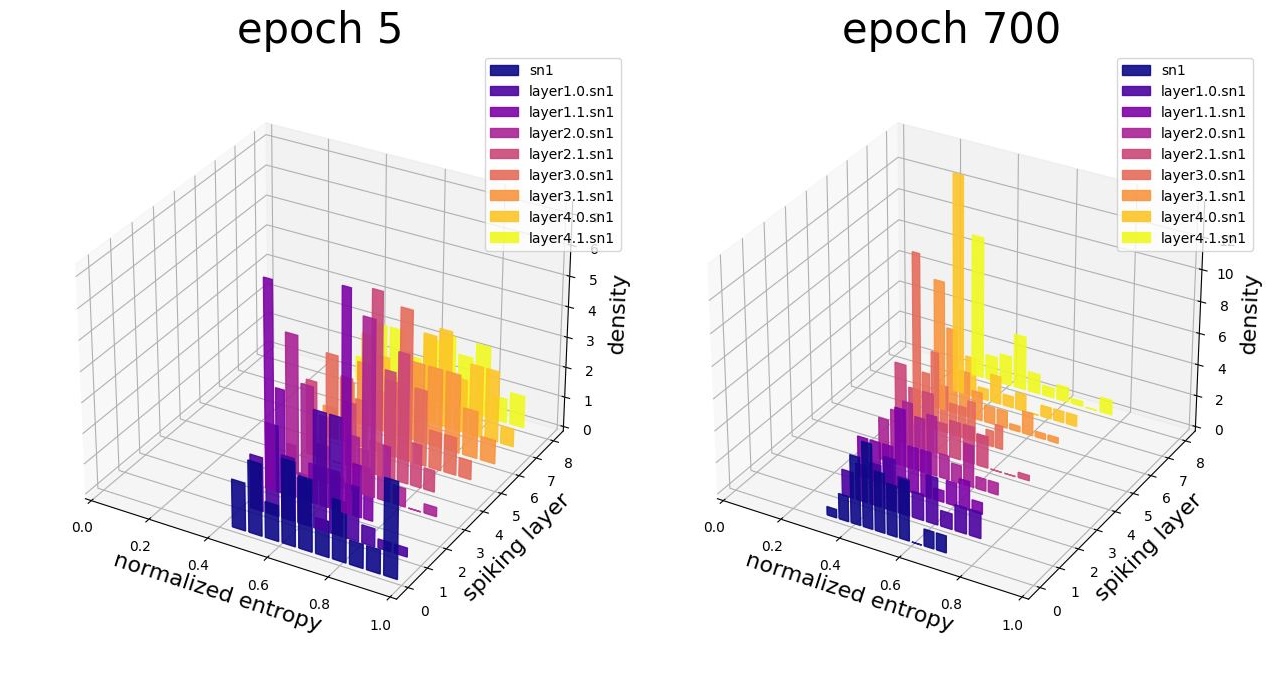}
\caption{Distribution of neurons by entropy of class prediction probabilities (lower entropy = more class-specialized). \textbf{Left}: early stage of learning (epoch \#5), \textbf{right}: late stage of learning (epoch \#700).}
\label{fig:layer_spec}
\end{figure}

\OK{
The results are shown in Fig.\thinspace\ref{fig:layer_spec}. One can see a sharp peak of low-entropy MEIs in the last spiking layers of the trained model, which corresponds to high model confidence in assigning specific classes to MEIs. Even in the middle of the network, most neurons are already specifically associated with a certain class. In contrast, at the early stage of training, the model does not show any MEI specificity for a particular class.
}

\begin{figure}[H]
\centering
\includegraphics[width=0.8\textwidth]{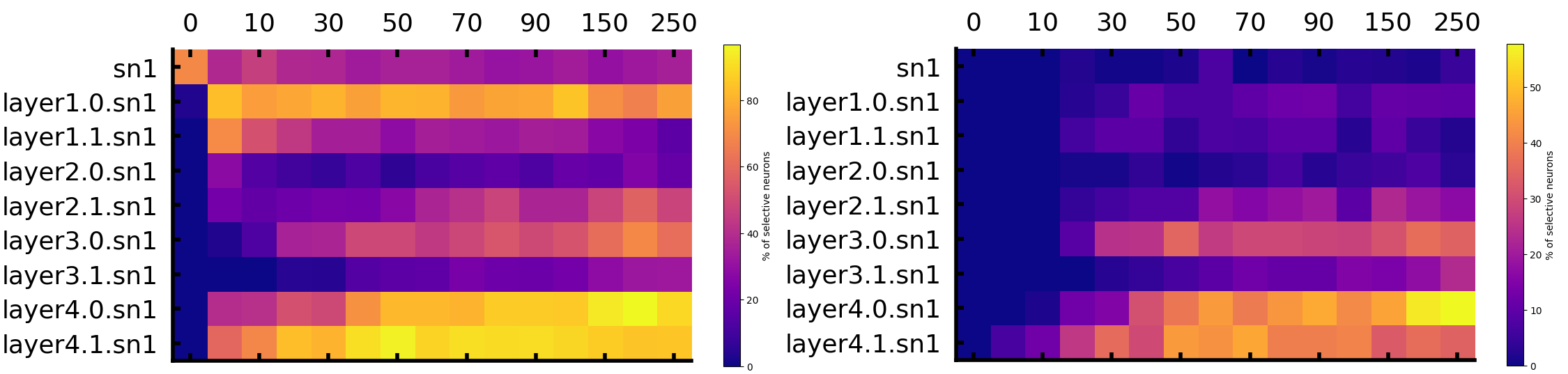}
\caption{{Fraction of selective neurons. Horizontal axis - epoch number, vertical axis - layers (lower = deeper). \textbf{Left}: activation threshold – spikes $\geq$75\% of exposure time; \textbf{Right:} additionally, max class probability $\geq$75\%, with max class stable  over different maximization methods.}}
\label{fig:fraction-of-selective-neurons}
\end{figure}

\OK{Notably, the three TT-based methods used often converged to similar images. Given the high complexity of the underlying landscape of potential MEIs, we used the following composite criterion of neuron selectivity with respect to a specific class:}
\begin{itemize}
    \itemsep0em 
    \item ``Stability'': All three maximization techniques converged on images of the same class (the MEI class was defined by the maximum probability at the network's output).
    \item ``Confidence'': The probability of the best class was at least 75\%.
    \item ``Activation'': The activity of the targeted neuron was at least 75\% of the maximum possible. Here, activity was calculated by simply adding up the number of spikes across all channels for a given neuron.
\end{itemize}

\OK{
As expected, the vast majority of specialized neurons were found in the final layers of the model, where MEIs contained the most complex patterns (Fig. \ref{fig:fraction-of-selective-neurons}, see Appendix \ref{appendix:MEI-gallery} for characteristic examples of MEIs). Surprisingly, however, we also found a significant number of neurons related to a certain class in the intermediate layers, accounting for up to 40\% of the total. Unexpectedly, highly selective neurons were found as early as the third spiking layer (Fig.\thinspace\ref{fig:argmax-pink-horse}).
}

\begin{figure}[H]
\centering
\includegraphics[width=0.8\textwidth]{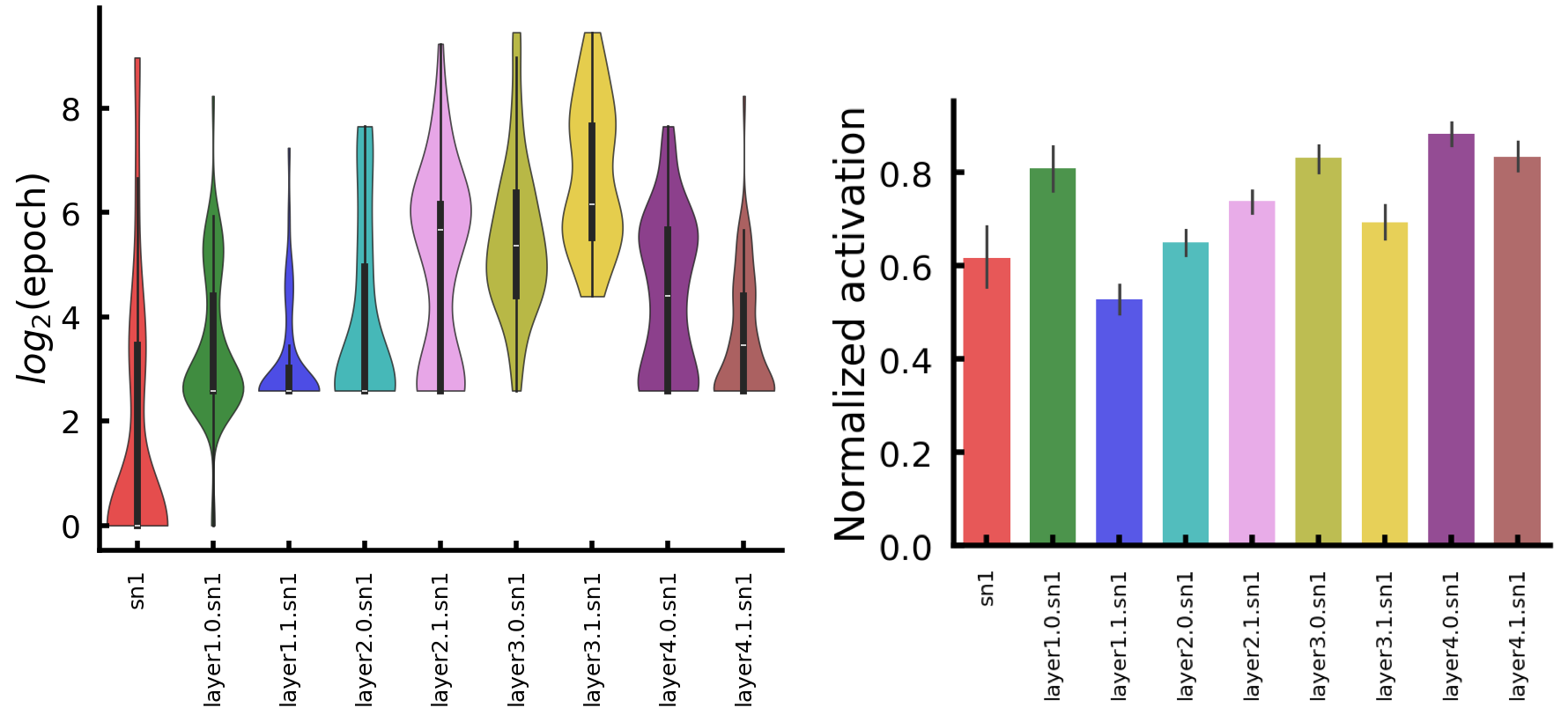}
\caption{
Emergence of neuronal specializations. Horizontal axis – layer depth (increasing left to right). \textbf{Left}: Distribution (over several runs) of the first epoch numbers when a strong MEI appears (neuron spikes $\geq$75\% of exposure time). \textbf{Right}: Normalized activation by MEI (errorbars – 95\% confidence interval).}
\label{fig:image-acts}
\end{figure}

\begin{figure}[H]
\centering
\includegraphics[width=0.8\textwidth]{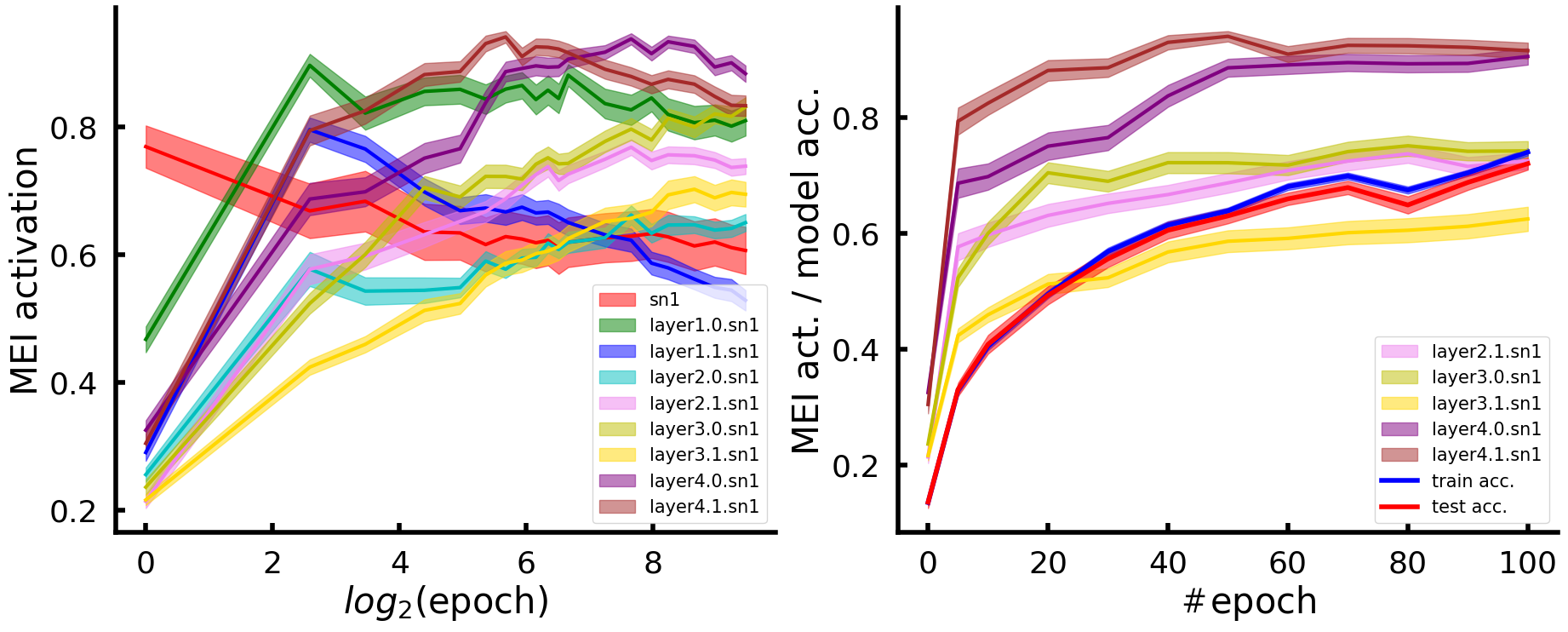}
\caption{{\textbf{Left}: Dynamics of neurons' activations in response to their MEIs. \textbf{Right}: the same for the first 100 epochs with the model classification accuracy in the same axes.}}
\label{fig:activation-dynamics}
\end{figure}

\OK{
As the model learns, highly specialized neurons increase in number. To quantify the ``goodness'' of MEIs, we calculated averaged neuronal response caused by MEIs across layers and learning epochs. The results are shown in Figures \ref{fig:image-acts} and \ref{fig:activation-dynamics}. In general, MEI-related activity increases with learning, correlating well with model performance (Figure \ref{fig:activation-dynamics} right). However, we observed an abnormally late development of MEIs before the final layers of the neural network (layer 3.1). These MEIs led to weaker activation of targeted neurons (Figure \ref{fig:image-acts} right). We believe this decrease may be due to ongoing restructuring of MEIs in these layers, as they act as a "gateway" to more complex specialized layers. It is also possible that insufficient optimization budget prevented obtaining the ``true'' MEIs of these neurons.
}

\begin{figure}[H]
\centering
\includegraphics[width=1.0\textwidth]{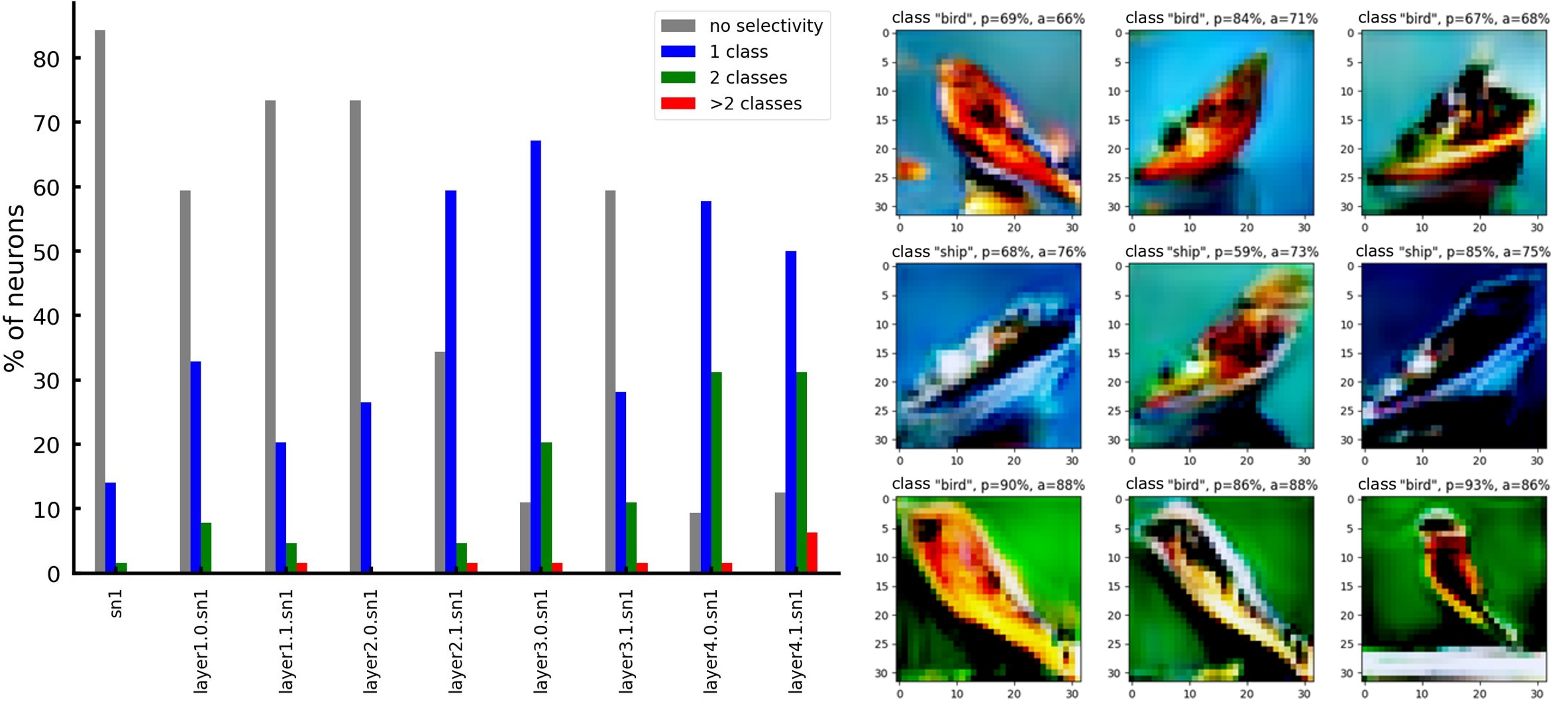}
\caption{\textbf{Left:} fractions of non-specialized, stable (selective to only 1 class) and labile (selective to 2 or more) neurons  vs layer depth. \textbf{Right:} oscillating MEIs of labile \textbf{\textit{``bird-ship'' neuron}}, \textbf{rows}: epoch \# 40, 100, 300 of training; \textbf{columns:} results obtained with TT, TT-S, TT-B optimization methods. The captions above each MEI provide information about the predominant class according to classification results, the probability, and the activation of the neuron as a percentage of the maximum possible.
}
\label{fig:labile-neurons-distr-and-bird-ship}
\end{figure}

\OK{
We then addressed the issue of ``labile'' neurons, which change their selectivity as training progresses. To do this, we tracked neurons that met the criteria for specialization on more than one class during training. These neurons are interesting because they help understand the mechanisms of ``narrow'' specializations formation. These often specialize on abstract concepts that can be used as building blocks for classifying more complex patterns. An example of MEI of such a neuron can be seen on Fig.\thinspace\ref{fig:fraction-of-selective-neurons}, right. It selectively activates on elongated, diagonally arranged structures on a blue/green background. In this area of the latent space, the embeddings of birds and ships are similar, which is reflected in the neuron's classification at different stages of training. Such labile neurons were virtually non-existent in the initial layers of the network (see Fig.\thinspace\ref{fig:fraction-of-selective-neurons}, left), but their fraction rose to approximately 40\% in the final layer. This could indicate a high degree of variability in the specialization of the final layers and their capacity for rapid reconfiguration when receiving new training data, drawing on information from the neurons in earlier layers. Specifically, this supports the empirical principle behind reservoir computing, which is to train only the output layers while allowing the internal layers to identify complex features in the data set \cite{Tanaka2019}.
}

\subsection{Complexity and diversity of specializations}\label{subsec:complexity-and-diversity}

\OK{Based on our observations, the entropy of class probability predictions by the model on MEIs decreases with layer depth and training epochs number, indicating that the neurons are gradually forming more refined class-related specializations.}

\OK{In general, as expected, neurons of early layers are maximally activated by simple geometric features or color patterns, while neurons in deeper layers show selective activation for one class or another (see Figs.\thinspace\ref{fig:MEI-gallery-0}, \ref{fig:MEI-gallery-1}, \ref{fig:MEI-gallery-2} for examples).}

\OK{MEIs generated using different optimization methods contain the same patterns of varying degrees of complexity, but are not identical to each other. In order to evaluate the variability of the resulting images, we calculated the average Euclidean and cosine distances between the latent vectors corresponding to MEIs (SN-GAN generator was used).}

\begin{figure}[H]
\centering
\includegraphics[width=1.0\textwidth]{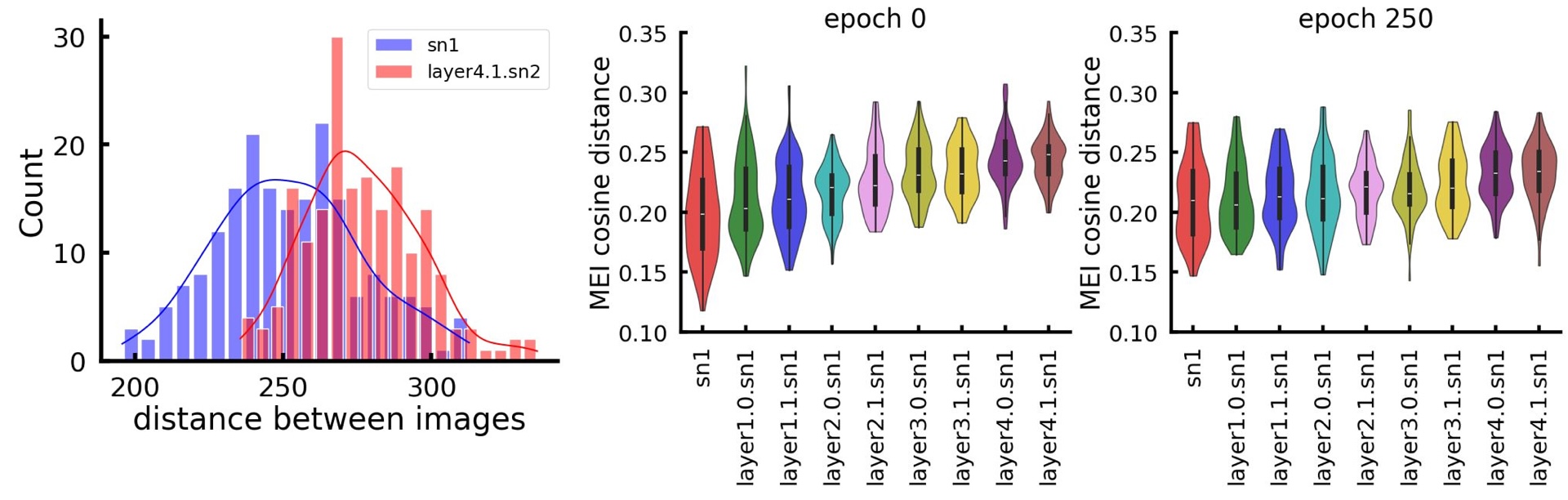}
\caption{Diversity of specializations in terms of distances between MEIs in the latent space. \textbf{Left}: Euclidean distances. \textbf{Right}: cosine similarities for different layers (layer depth increases to the right), at different training stages.}
\label{fig:image-dists}
\end{figure}

\OK{The variability of possible images in the last layers of spiking ResNet18 turned out to be greater than in the first ones (see Fig.~\ref{fig:image-dists}). Average distances between the generated MEIs in the last spiking layer turned out to be about 20\% higher than in the first one, revealing significant transformations of neurons' specializations landscape. Possible explanation of this effect is that there are many ways in which a complex concept, such as a horse, can be realised in an image. NNs learn to nonlinearly project the data space during training, simultaneously identifying some of its areas with each other \cite{nnmap}. Thus, the number of local maxima in the overall image space grows with increasing specialization of neurons.
At the same time, this effect measured via cosine distance became weaker during model training (see Fig.\thinspace\ref{fig:image-dists}, right).
}

\OK{The growing diversity of MEIs is accompanied by an increase in the activations they produce on corresponding neurons(~\ref{fig:image-acts}). This may be explained by finer specialization of neurons in deeper layers, in accordance with the results from section \ref{subsec:emergence-and-dynamics}. At the same time, the dips in maximal activation values observed in layer 3.1 are reproduced by both generator types (GAN and VAE) and show that the overall picture may be more complex.}

\begin{figure}[H]
\centering
\includegraphics[width=0.75\textwidth]{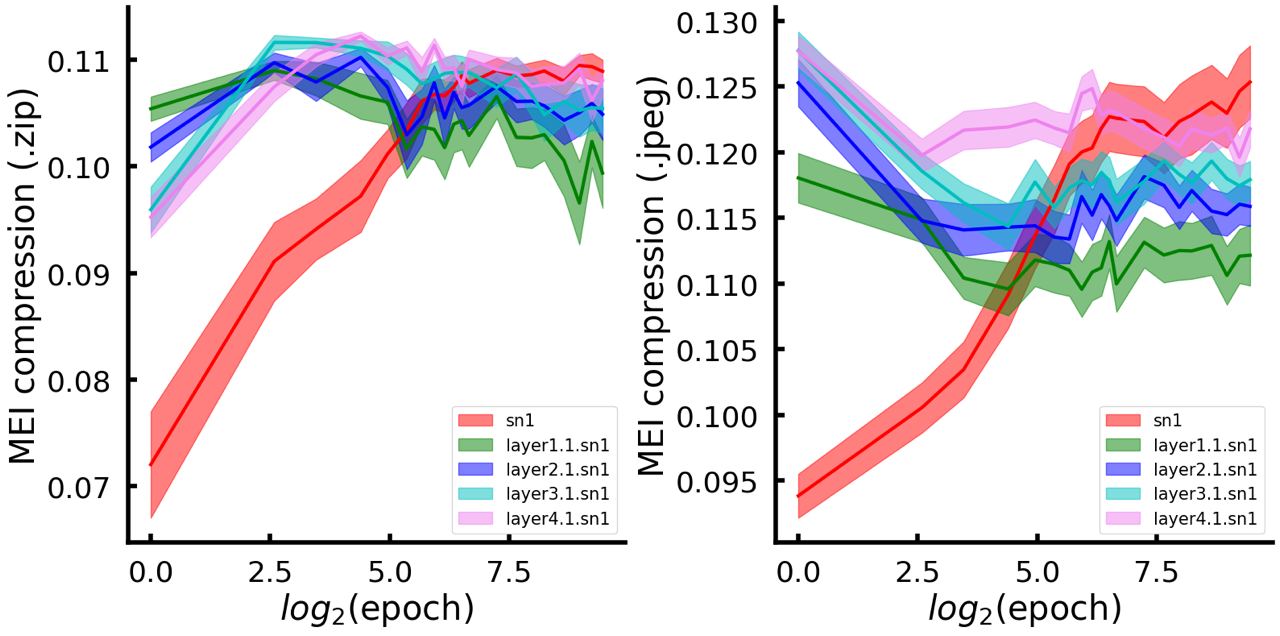}
\caption{Complexity of MEIs, measured as compressed file size (JPEG and ZIP compressions).}
\label{fig:complexity-by-compression}
\end{figure}

\OK{As the network learns, the content of MEIs becomes more complex. Calculating an exact measure of image complexity is a challenging task, for which several complementary approaches have been developed \cite{Chikhman2012}. However, it is possible to gain a general understanding of this measure by using simple heuristics based on information compression algorithms. This approach has proven successful in assessing entropy and mutual information \cite{Baronchelli2005} as well as the quantitative assessment of consciousness level \cite{Casali2013}. To quantify visual complexity of generated MEIs, we calculated the compression ratios of the optimized images in the \texttt{.jpg} and \texttt{.zip} formats and compared them to the size of the original images (see Fig.\thinspace\ref{fig:complexity-by-compression}). Both compression methods yielded similar results: as the network learns, the complexity of MEIs in the first layers increases, gradually reaching a steady state. We believe that this is due to rapid formation of more complex specializations in the early layers of the network (Fig.\thinspace\ref{fig:image-acts} see section \ref{subsec:neuronal-selectivity} in Discussion).}

\section{Discussion}\label{sec:Discussion}

\subsection{Neuronal selectivity in spiking convolutional networks}\label{subsec:neuronal-selectivity}

\OK{
We track neuronal selectivity properties throughout the training process of the model and observe MEIs of varying complexity with more refined features corresponding to deeper layers, in accordance with the principles shown in both ANNs \cite{Bengio-transferable-features,AlexNet,erhan2009visualizing} and animal visual cortex \cite{DiCarlo2012, QuianQuiroga2023}.
}

\OK{
Neurons from intermediate layers are often selective for ambiguous images potentially useful for multiple class discrimination (see Fig.\thinspace\ref{fig:MEI-gallery-1} (left) for examples). This effect can provide a potential basis for complex image specializations. For instance, it has been shown that abstract feature representations emerge naturally in neural networks trained to perform multiple tasks \cite{Johnston2023}.
}

\OK{
The influence of different sensory modalities on the visual cortex in animals is not negligible, potentially leading to complex neuron specializations \cite{Pennartz2023}. For example, neurons in the V2 visual area can be selective for ``naturalistic'' texture stimuli \cite{Movshon2014}. Neurons selective for complex vocal stimuli have also been found in the primary auditory cortex \cite{MontesLourido2021}. These findings coincide with the formation of complex specializations in early layers, as demonstrated in this work (see Fig.\thinspace\ref{fig:fraction-of-selective-neurons}).
}

\OK{
It is notable that neuronal specializations in our setting form early during the training process, as shown in Fig.\thinspace\ref{fig:fraction-of-selective-neurons} – this phenomenon has also been observed in non-spiking ANNs \cite{critical-learning-periods}. The model's neurons are able to form complex representations, expressed in the activity of neurons, after several exposures to the dataset. This occurs both in deep and intermediate layers, even though the recognition quality has not yet reached a plateau at this stage of learning. This is consistent with data on the early formation of specialized neurons in biological neural networks, as obtained in experiments \cite{Sotskov2022}. This behavior may indicate that complex neural representations are rapidly formed, and the network then ``rebuilds'' the weights over time to optimally convey information from these highly specialized neurons to the output layers.
}

The results obtained suggest that the reconfiguration of neural specialization is a fundamental process in both biological and artificial neural networks. Additionally, heterogeneity in the selectivity of neurons is observed in these systems, with some neurons retaining their specialization as they learn, while others' specializations are more labile. In biological networks, it was shown that neurons with stronger selectivity tend to be more stable, supporting the hypothesis of the log-dynamic brain \cite{log_dynamic}. In future work, we plan to investigate how this behavior relates to the strength of specialization, as well as investigate the weights of connections between artificial neurons with specific specializations in order to gain better understanding of their influence on one another and trace the development of complex specializations.

\subsection{Limitations of the current work and directions of future research}\label{subsec:limitations}

As our results show, the quality of the calculated MEIs depends significantly on the image generator used. In the future, we plan to extend MANGO by adding support for new generators that are well suited for the MEI search task (i.e. combining the discrete structure of the latent space with the high plausibility of the generated images). It is of interest to consider the popular direction of discrete GANs \cite{Taming-transformers}.

An important research question remains on the patterns and mechanisms of neuronal specializations formation over time. Although in present work we examined MEIs for an already trained network, it is of great interest to be able to monitor their evolution during the training process. We plan to add this feature to MANGO framework in the future.

One of the most interesting and potentially fruitful applications of our tools is maximizing neuronal activation \textit{in vivo}. Since biological neurons have an irreducible minimum response time to a stimulus, research into algorithms aimed at reducing optimizer budget required for MEI generation seems particularly relevant. In particular, we plan to extend MANGO with more optimization methods based on Tensor Train decomposition, specifically optimized for this task.

In parallel with single-neuron selectivity, neuronal computations can be described by low-dimensional activity manifolds, which arise from the collective, coordinated spiking of numerous neurons. This type of information encoding in the brain is known as population coding and has a long and rich history in neuroscience \cite{Panzeri2015, Gallego2017}. Recently, population behavior of artificial neurons in large language models (LLMs) has received particular attention, potentially leading to the emergence of the new field of representation engineering \cite{repE}. To our knowledge, the relationships between optimal stimuli for single neurons and their contribution to population coding have not yet been analyzed, and we consider this to be one of the important directions for future work.

Recently, it has become apparent that the activity of individual components can significantly influence the operation of a large neural network. Researchers working with LLMs have recently demonstrated that manipulating the activity of a single specialized artificial neuron can lead to significant changes in the behavior of an LLM \cite{templeton2024scaling}. At the same time, it is known that stimulating a small number of neurons (or even a single neuron in some cases) in a biological neural network can cause the entire ensemble to become highly active, potentially influencing animal behavior \cite{AlejandreGarca2022}. Given this evidence of the importance of individual computational units, it would be interesting to investigate the effect of inhibiting or exciting a single neuron on model performance and correlate this with the MEI properties of that neuron. We plan to conduct such analysis in future work.

%\TODO{transfer learning}

%\TODO{Layerwise AM}.

\subsection{Activation maximization in ANNs \& explainable AI}\label{subsec:explainable-AI}

\OK{The problem of maximizing neuronal activation in artificial neural networks (ANNs) is being actively researched within the field of ANN visualization \& interpretation \cite{erhan2009visualizing,samek2016-evaluating-visualizations-of-what-NN-learns,DNN-visualization-Yosinski,methods-for-interpreting-NNs}, a sub-field of the more general area of explainable AI. In order to successfully integrate ANN-based solutions into critical systems, such as medical and legal practices, it is necessary to have a human-understandable interpretation of the algorithm's decision-making process. In recent years, methods based on visualizing the computation graph, loss function, parameter space of certain layers, or even individual neurons \cite{Bau-2020-role-of-individual-neurons} have gained popularity in creating interpretable deep learning models. These methods allow researchers to better understand the inner workings of NNs.}

\OK{A recent review \cite{matveev2021overview} of visualisation techniques for unravelling neuronal stimuli covers some of the most successful approaches: the method of activation maximization \cite{erhan2009visualizing}, the Grad-CAM method \cite{selvaraju2017grad}, layer-wise relevance propagation (LRP) \cite{methods-for-interpreting-NNs} technique, well as the Integrated Gradients method \cite{sundararajan2017axiomatic}. Another outstanding recent work is \cite{goh2021multimodal}, which demonstrates a rich range of stimuli that maximise the activation of different neurons in an ANN. This work also demonstrates the effect of multimodal selectivity in artificial neurons.}

\OK{However, all of the aforementioned activation maximization methods are based on automatic differentiation: the ANN's computation graph is formed allowing to compute gradients with minimal computational complexity. This allows the use of gradient descent (ascent), and its stochastic modifications. Gradient-based methods are clearly not applicable to living systems, so one must either use gradient-free methods, such as genetic algorithms (as in \cite{XDream}), or other methods that use local approximations of gradients, such as evolutionary algorithms, or alternative approaches, including optimization techniques based on TT-decomposition.} 

%Notably genetic and evolutionary algorithms were considered in \cite{wang2022high} for activation maximization in ANNs.

\OK{Our results presented in section \ref{subsec:performance} indicate that TT-based optimization methods outperform Nevergrad's \cite{Nevergrad} Portfolio in our problem setting. Portfolio is a balanced mixture of optimization methods that includes CMA-ES \cite{CMA}. It was shown \cite{wang2022high} that CMA-ES outperforms other genetic algorithms on the problem of activation maximization. This allows one to carefully conclude that TT-based optimization methods are more suitable for AM problems, although this claim requires further accurate examination.}

\subsection{Possible applications for TT-based activation maximization in biological neurons}\label{subsec:in-vivo}

\OK{Modern optical imaging techniques allow us to monitor the activity of hundreds of neurons in their temporal dynamics in vivo. Activation maximization of multiple neurons, both individually and in groups, could be a powerful tool for analyzing the function of the nervous system at the cellular level, shedding light on the formation and spread of cognitive specializations in living neurons. However, the design of biological experiments poses serious limitations on the applicability of optimization methods. These methods need to be quick, accurate, and able to tune themselves to new data. We expect our TT-decomposition-based methods to possess all these properties.}

\OK{A separate challenge is finding optimal stimuli in non-visual domains, such as the space of sounds or smells, which may be far more relevant to the animal under study than to humans \cite{Arakawa2008}. We also hope to gain some insight into the applicability of TT-decomposition based methods in these latent stimulus spaces.}

\OK{Studies of neuronal selectivity in spiking neural networks are of great interest, as they reveal the mechanisms of the functional units' operation in the neural network using a biologically plausible model. The use of spiking artificial neurons makes the analysis in this study significantly closer to real electrophysiological experiments \cite{ponce2019evolving} and allows us to test hypotheses about the temporal patterns of neurons' responses to MEIs in silico. We hope that our algorithms will be useful in the future study of optimal stimuli for biological neural networks.}

\OK{
Our framework aims to bridge the gap between neurobiological and computational experiments, enabling the possibility to explore the principles of information encoding in deep neural networks. Additionally, we hope that this research will bring activation maximization algorithms closer to the requirements of real biological experiments, providing a new tool for analyzing cognitive functions of individual neurons.}

\section{Acknowledgements} 
\OK{
This work was supported by the Non-Commercial Foundation for the Support of Science and Education ``INTELLECT'' and Lomonosov Moscow State University. N. Pospelov acknowledges the support of the Brain Program at the IDEAS Research Center. The main computational experiments were conducted on the ABS hardware. Some experiments used the computational resources of the HPC facilities at HSE University \cite{HSE-Supercomputer-counter-paper}. The authors are grateful to all members of the Laboratory of Neuronal Intelligence and especially to Ksenia A. Toropova for fruitful discussions. 
}

\section{References} 
%	\newpage
\bibliography{refs}

\begin{thebibliography}{10}

\bibitem{HubelWiesel}
D.~H. Hubel and T.~N. Wiesel, {\em Brain and visual perception: the story of a 25-year collaboration}.
\newblock Oxford University Press, 2004.

\bibitem{BiPOLES}
J.~Vierock, S.~Rodriguez-Rozada, A.~Dieter, F.~Pieper, R.~Sims, F.~Tenedini, A.~C. Bergs, I.~Bendifallah, F.~Zhou, N.~Zeitzschel, {\em et~al.}, ``Bipoles is an optogenetic tool developed for bidirectional dual-color control of neurons,'' {\em Nature communications}, vol.~12, no.~1, p.~4527, 2021.

\bibitem{QuianQuiroga2023}
R.~Quian~Quiroga, M.~Boscaglia, J.~Jonas, H.~G. Rey, X.~Yan, L.~Maillard, S.~Colnat-Coulbois, L.~Koessler, and B.~Rossion, ``Single neuron responses underlying face recognition in the human midfusiform face-selective cortex,'' {\em Nature Communications}, vol.~14, Sept. 2023.

\bibitem{ponce2019evolving}
C.~R. Ponce, W.~Xiao, P.~F. Schade, T.~S. Hartmann, G.~Kreiman, and M.~S. Livingstone, ``Evolving images for visual neurons using a deep generative network reveals coding principles and neuronal preferences,'' {\em Cell}, vol.~177, no.~4, pp.~999--1009, 2019.

\bibitem{bardon2022face}
A.~Bardon, W.~Xiao, C.~R. Ponce, M.~S. Livingstone, and G.~Kreiman, ``Face neurons encode nonsemantic features,'' {\em Proceedings of the national academy of sciences}, vol.~119, no.~16, p.~e2118705119, 2022.

\bibitem{neuroethics2019}
A.~J. Shriver and T.~M. John, ``Neuroethics and animals: report and recommendations from the university of pennsylvania animal research neuroethics workshop,'' {\em ILAR journal}, vol.~60, no.~3, pp.~424--433, 2019.

\bibitem{neuroethics2023}
P.~Singer and Y.~F. Tse, ``Ai ethics: the case for including animals,'' {\em AI and Ethics}, vol.~3, no.~2, pp.~539--551, 2023.

\bibitem{Brain-is-not-like-ANN}
A.~M. Zador, ``A critique of pure learning and what artificial neural networks can learn from animal brains,'' {\em Nature communications}, vol.~10, no.~1, p.~3770, 2019.

\bibitem{Brain-is-not-like-ANN-2}
R.~Schaeffer, M.~Khona, and I.~Fiete, ``No free lunch from deep learning in neuroscience: A case study through models of the entorhinal-hippocampal circuit,'' {\em Advances in Neural Information Processing Systems}, vol.~35, pp.~16052--16067, 2022.

\bibitem{ANN-neuron-specializations}
K.~Dobs, J.~Martinez, A.~J. Kell, and N.~Kanwisher, ``Brain-like functional specialization emerges spontaneously in deep neural networks,'' {\em Science advances}, vol.~8, no.~11, p.~eabl8913, 2022.

\bibitem{Multimodal-neurons-in-ANNs}
G.~Goh, N.~Cammarata, C.~Voss, S.~Carter, M.~Petrov, L.~Schubert, A.~Radford, and C.~Olah, ``Multimodal neurons in artificial neural networks,'' {\em Distill}, vol.~6, no.~3, p.~e30, 2021.

\bibitem{bau2017network}
D.~Bau, B.~Zhou, A.~Khosla, A.~Oliva, and A.~Torralba, ``Network dissection: Quantifying interpretability of deep visual representations,'' in {\em Proceedings of the IEEE conference on computer vision and pattern recognition}, pp.~6541--6549, 2017.

\bibitem{Olah-feature-visualization}
C.~Olah, A.~Mordvintsev, and L.~Schubert, ``Feature visualization,'' {\em Distill}, 2017.
\newblock https://distill.pub/2017/feature-visualization.

\bibitem{Nguyen2019}
A.~Nguyen, J.~Yosinski, and J.~Clune, {\em Understanding Neural Networks via Feature Visualization: A Survey}, p.~55–76.
\newblock Springer International Publishing, 2019.

\bibitem{wang2022high}
B.~Wang and C.~R. Ponce, ``High-performance evolutionary algorithms for online neuron control,'' in {\em Proceedings of the Genetic and Evolutionary Computation Conference}, pp.~1308--1316, 2022.

\bibitem{GANpaper}
I.~Goodfellow, J.~Pouget-Abadie, M.~Mirza, B.~Xu, D.~Warde-Farley, S.~Ozair, A.~Courville, and Y.~Bengio, ``Generative adversarial nets,'' {\em Advances in neural information processing systems}, vol.~27, 2014.

\bibitem{TTOpt-paper}
K.~Sozykin, A.~Chertkov, R.~Schutski, A.-H. Phan, A.~S. CICHOCKI, and I.~Oseledets, ``Ttopt: A maximum volume quantized tensor train-based optimization and its application to reinforcement learning,'' {\em Advances in Neural Information Processing Systems}, vol.~35, pp.~26052--26065, 2022.

\bibitem{TT-Oseledets}
I.~V. Oseledets, ``Tensor-train decomposition,'' {\em SIAM Journal on Scientific Computing}, vol.~33, no.~5, pp.~2295--2317, 2011.

\bibitem{SNNs-what-are-overview}
K.~Yamazaki, V.-K. Vo-Ho, D.~Bulsara, and N.~Le, ``Spiking neural networks and their applications: A review,'' {\em Brain Sciences}, vol.~12, no.~7, p.~863, 2022.

\bibitem{Panzeri2001}
S.~Panzeri, R.~S. Petersen, S.~R. Schultz, M.~Lebedev, and M.~E. Diamond, ``The role of spike timing in the coding of stimulus location in rat somatosensory cortex,'' {\em Neuron}, vol.~29, p.~769–777, Mar. 2001.

\bibitem{AndradeTalavera2023}
Y.~Andrade-Talavera, A.~Fisahn, and A.~Rodríguez-Moreno, ``Timing to be precise? an overview of spike timing-dependent plasticity, brain rhythmicity, and glial cells interplay within neuronal circuits,'' {\em Molecular Psychiatry}, vol.~28, p.~2177–2188, Mar. 2023.

\bibitem{SNN-neural-representations-2023}
Y.~Li, Y.~Kim, H.~Park, and P.~Panda, ``Uncovering the representation of spiking neural networks trained with surrogate gradient,'' {\em arXiv preprint arXiv:2304.13098}, 2023.

\bibitem{Inception-what-excites-neurons-most}
E.~Y. Walker, F.~H. Sinz, E.~Cobos, T.~Muhammad, E.~Froudarakis, P.~G. Fahey, A.~S. Ecker, J.~Reimer, X.~Pitkow, and A.~S. Tolias, ``Inception loops discover what excites neurons most using deep predictive models,'' {\em Nature neuroscience}, vol.~22, no.~12, pp.~2060--2065, 2019.

\bibitem{erhan2009visualizing}
D.~Erhan, Y.~Bengio, A.~Courville, and P.~Vincent, ``Visualizing higher-layer features of a deep network,'' {\em University of Montreal}, vol.~1341, no.~3, p.~1, 2009.

\bibitem{DNN-visualization-Yosinski}
J.~Yosinski, J.~Clune, A.~Nguyen, T.~Fuchs, and H.~Lipson, ``Understanding neural networks through deep visualization,'' {\em arXiv preprint arXiv:1506.06579}, 2015.

\bibitem{samek2016-evaluating-visualizations-of-what-NN-learns}
W.~Samek, A.~Binder, G.~Montavon, S.~Lapuschkin, and K.-R. M{\"u}ller, ``Evaluating the visualization of what a deep neural network has learned,'' {\em IEEE transactions on neural networks and learning systems}, vol.~28, no.~11, pp.~2660--2673, 2016.

\bibitem{methods-for-interpreting-NNs}
G.~Montavon, W.~Samek, and K.-R. M{\"u}ller, ``Methods for interpreting and understanding deep neural networks,'' {\em Digital signal processing}, vol.~73, pp.~1--15, 2018.

\bibitem{Nevergrad}
J.~Rapin and O.~Teytaud, ``{Nevergrad - A gradient-free optimization platform}.'' \url{https://GitHub.com/FacebookResearch/Nevergrad}, 2018.

\bibitem{XDream}
W.~Xiao and G.~Kreiman, ``Xdream: Finding preferred stimuli for visual neurons using generative networks and gradient-free optimization,'' {\em PLoS computational biology}, vol.~16, no.~6, p.~e1007973, 2020.

\bibitem{sgd}
S.~Ruder, ``An overview of gradient descent optimization algorithms,'' 2016.

\bibitem{Gradient-free-AM}
W.~Xiao and G.~Kreiman, ``Gradient-free activation maximization for identifying effective stimuli,'' {\em arXiv preprint arXiv:1905.00378}, 2019.

\bibitem{CMA}
N.~Hansen and A.~Ostermeier, ``Adapting arbitrary normal mutation distributions in evolution strategies: The covariance matrix adaptation,'' in {\em Proceedings of IEEE international conference on evolutionary computation}, pp.~312--317, IEEE, 1996.

\bibitem{DL-Fathers-Nature-paper}
Y.~LeCun, Y.~Bengio, and G.~Hinton, ``Deep learning,'' {\em nature}, vol.~521, no.~7553, pp.~436--444, 2015.

\bibitem{DeepLearningBook}
I.~Goodfellow, Y.~Bengio, and A.~Courville, {\em Deep learning}.
\newblock MIT press, 2016.

\bibitem{Izhikevich-Spike-model}
E.~M. Izhikevich, ``Simple model of spiking neurons,'' {\em IEEE Transactions on neural networks}, vol.~14, no.~6, pp.~1569--1572, 2003.

\bibitem{Perceptron-McCulloch-Pitts}
W.~S. McCulloch and W.~Pitts, ``A logical calculus of the ideas immanent in nervous activity,'' {\em The bulletin of mathematical biophysics}, vol.~5, pp.~115--133, 1943.

\bibitem{Rosenblatt-Perceptron-1}
F.~Rosenblatt, {\em The perceptron, a perceiving and recognizing automaton Project Para}.
\newblock Cornell Aeronautical Laboratory, 1957.

\bibitem{Rosenblatt-Perceptron-2}
F.~Rosenblatt, ``The perceptron: a probabilistic model for information storage and organization in the brain.,'' {\em Psychological review}, vol.~65, no.~6, p.~386, 1958.

\bibitem{Backpropagation-Hinton}
D.~E. Rumelhart, G.~E. Hinton, and R.~J. Williams, ``Learning representations by back-propagating errors,'' {\em nature}, vol.~323, no.~6088, pp.~533--536, 1986.

\bibitem{Maass-SNN-origins}
W.~Maass, ``Networks of spiking neurons: the third generation of neural network models,'' {\em Neural networks}, vol.~10, no.~9, pp.~1659--1671, 1997.

\bibitem{SNN-Torch-paper}
J.~K. Eshraghian, M.~Ward, E.~Neftci, X.~Wang, G.~Lenz, G.~Dwivedi, M.~Bennamoun, D.~S. Jeong, and W.~D. Lu, ``Training spiking neural networks using lessons from deep learning,'' {\em Proceedings of the IEEE}, vol.~111, no.~9, pp.~1016--1054, 2023.

\bibitem{SpikingJelly-paper}
W.~Fang, Y.~Chen, J.~Ding, Z.~Yu, T.~Masquelier, D.~Chen, L.~Huang, H.~Zhou, G.~Li, and Y.~Tian, ``Spikingjelly: An open-source machine learning infrastructure platform for spike-based intelligence,'' {\em Science Advances}, vol.~9, no.~40, p.~eadi1480, 2023.

\bibitem{ResNet-paper}
K.~He, X.~Zhang, S.~Ren, and J.~Sun, ``Deep residual learning for image recognition,'' in {\em Proceedings of the IEEE conference on computer vision and pattern recognition}, pp.~770--778, 2016.

\bibitem{Hodgkin-Huxley-model}
A.~L. Hodgkin and A.~F. Huxley, ``A quantitative description of membrane current and its application to conduction and excitation in nerve,'' {\em The Journal of physiology}, vol.~117, no.~4, p.~500, 1952.

\bibitem{Leaky-Integrate-and-Fire-Lapicque}
N.~Brunel and M.~C. Van~Rossum, ``Lapicque’s 1907 paper: from frogs to integrate-and-fire,'' {\em Biological cybernetics}, vol.~97, no.~5-6, pp.~337--339, 2007.

\bibitem{Tensorizing-NNs}
A.~Novikov, D.~Podoprikhin, A.~Osokin, and D.~P. Vetrov, ``Tensorizing neural networks,'' {\em Advances in neural information processing systems}, vol.~28, 2015.

\bibitem{liu2022tt}
D.~Liu, L.~T. Yang, P.~Wang, R.~Zhao, and Q.~Zhang, ``Tt-tsvd: A multi-modal tensor train decomposition with its application in convolutional neural networks for smart healthcare,'' {\em ACM Transactions on Multimedia Computing, Communications, and Applications (TOMM)}, vol.~18, no.~1s, pp.~1--17, 2022.

\bibitem{Optima-TT-Chertkov}
A.~Chertkov, G.~Ryzhakov, G.~Novikov, and I.~Oseledets, ``Optimization of functions given in the tensor train format,'' {\em arXiv preprint arXiv:2209.14808}, 2022.

\bibitem{Cichocki-tensors-for-optimization-1}
A.~Cichocki, N.~Lee, I.~Oseledets, A.-H. Phan, Q.~Zhao, D.~P. Mandic, {\em et~al.}, ``Tensor networks for dimensionality reduction and large-scale optimization: Part 1 low-rank tensor decompositions,'' {\em Foundations and Trends{\textregistered} in Machine Learning}, vol.~9, no.~4-5, pp.~249--429, 2016.

\bibitem{Cichocki-tensors-for-optimization-2}
A.~Cichocki, A.-H. Phan, Q.~Zhao, N.~Lee, I.~Oseledets, M.~Sugiyama, D.~P. Mandic, {\em et~al.}, ``Tensor networks for dimensionality reduction and large-scale optimization: Part 2 applications and future perspectives,'' {\em Foundations and Trends{\textregistered} in Machine Learning}, vol.~9, no.~6, pp.~431--673, 2017.

\bibitem{PROTES-Chertkov-paper}
A.~Batsheva, A.~Chertkov, G.~Ryzhakov, and I.~Oseledets, ``{PROTES}: Probabilistic optimization with tensor sampling,'' {\em Advances in Neural Information Processing Systems}, 2023.

\bibitem{Simulated-annealing-what-is}
M.~W. Trosset, ``What is simulated annealing?,'' {\em Optimization and Engineering}, vol.~2, pp.~201--213, 2001.

\bibitem{REINFORCE-trick}
R.~J. Williams, ``Simple statistical gradient-following algorithms for connectionist reinforcement learning,'' {\em Machine learning}, vol.~8, pp.~229--256, 1992.

\bibitem{CIFAR10-dataset-reference}
A.~Krizhevsky, G.~Hinton, {\em et~al.}, ``Learning multiple layers of features from tiny images,'' 2009.

\bibitem{SN-GAN-paper}
T.~Miyato, T.~Kataoka, M.~Koyama, and Y.~Yoshida, ``Spectral normalization for generative adversarial networks,'' {\em arXiv preprint arXiv:1802.05957}, 2018.

\bibitem{VAE-vanilla-Kingma-Welling}
D.~P. Kingma and M.~Welling, ``Auto-encoding variational bayes,'' {\em arXiv preprint arXiv:1312.6114}, 2013.

\bibitem{VQ-VAE-Van-Den-Oord}
A.~Van Den~Oord, O.~Vinyals, {\em et~al.}, ``Neural discrete representation learning,'' {\em Advances in neural information processing systems}, vol.~30, 2017.

\bibitem{GAN-metric-inception-score}
T.~Salimans, I.~Goodfellow, W.~Zaremba, V.~Cheung, A.~Radford, and X.~Chen, ``Improved techniques for training gans,'' {\em Advances in neural information processing systems}, vol.~29, 2016.

\bibitem{GAN-metric-Frechet}
M.~Heusel, H.~Ramsauer, T.~Unterthiner, B.~Nessler, and S.~Hochreiter, ``Gans trained by a two time-scale update rule converge to a local nash equilibrium,'' {\em Advances in neural information processing systems}, vol.~30, 2017.

\bibitem{MNIST}
Y.~LeCun, ``The mnist database of handwritten digits,'' {\em http://yann. lecun. com/exdb/mnist/}, 1998.

\bibitem{Fashion-MNIST}
H.~Xiao, K.~Rasul, and R.~Vollgraf, ``Fashion-mnist: a novel image dataset for benchmarking machine learning algorithms,'' {\em arXiv preprint arXiv:1708.07747}, 2017.

\bibitem{Imagenet}
J.~Deng, W.~Dong, R.~Socher, L.-J. Li, K.~Li, and L.~Fei-Fei, ``Imagenet: A large-scale hierarchical image database,'' in {\em 2009 IEEE conference on computer vision and pattern recognition}, pp.~248--255, Ieee, 2009.

\bibitem{AlexNet}
A.~Krizhevsky, I.~Sutskever, and G.~E. Hinton, ``Imagenet classification with deep convolutional neural networks,'' {\em Advances in neural information processing systems}, vol.~25, 2012.

\bibitem{Densenet}
G.~Huang, Z.~Liu, L.~Van Der~Maaten, and K.~Q. Weinberger, ``Densely connected convolutional networks,'' in {\em Proceedings of the IEEE conference on computer vision and pattern recognition}, pp.~4700--4708, 2017.

\bibitem{VGG=VeryDeepCNN}
K.~Simonyan and A.~Zisserman, ``Very deep convolutional networks for large-scale image recognition,'' {\em arXiv preprint arXiv:1409.1556}, 2014.

\bibitem{Tanaka2019}
G.~Tanaka, T.~Yamane, J.~B. Héroux, R.~Nakane, N.~Kanazawa, S.~Takeda, H.~Numata, D.~Nakano, and A.~Hirose, ``Recent advances in physical reservoir computing: A review,'' {\em Neural Networks}, vol.~115, p.~100–123, July 2019.

\bibitem{nnmap}
G.~Mont\'{u}far, R.~Pascanu, K.~Cho, and Y.~Bengio, ``On the number of linear regions of deep neural networks,'' in {\em Proceedings of the 27th International Conference on Neural Information Processing Systems - Volume 2}, NIPS'14, (Cambridge, MA, USA), p.~2924–2932, MIT Press, 2014.

\bibitem{Chikhman2012}
V.~Chikhman, V.~Bondarko, M.~Danilova, A.~Goluzina, and Y.~Shelepin, ``Complexity of images: Experimental and computational estimates compared,'' {\em Perception}, vol.~41, p.~631–647, Jan. 2012.

\bibitem{Baronchelli2005}
A.~Baronchelli, E.~Caglioti, and V.~Loreto, ``Measuring complexity with zippers,'' {\em European Journal of Physics}, vol.~26, p.~S69–S77, July 2005.

\bibitem{Casali2013}
A.~G. Casali, O.~Gosseries, M.~Rosanova, M.~Boly, S.~Sarasso, K.~R. Casali, S.~Casarotto, M.-A. Bruno, S.~Laureys, G.~Tononi, and M.~Massimini, ``A theoretically based index of consciousness independent of sensory processing and behavior,'' {\em Science Translational Medicine}, vol.~5, Aug. 2013.

\bibitem{Bengio-transferable-features}
J.~Yosinski, J.~Clune, Y.~Bengio, and H.~Lipson, ``How transferable are features in deep neural networks?,'' {\em Advances in neural information processing systems}, vol.~27, 2014.

\bibitem{DiCarlo2012}
J.~J. DiCarlo, D.~Zoccolan, and N.~C. Rust, ``How does the brain solve visual object recognition?,'' {\em Neuron}, vol.~73, p.~415–434, Feb. 2012.

\bibitem{Johnston2023}
W.~J. Johnston and S.~Fusi, ``Abstract representations emerge naturally in neural networks trained to perform multiple tasks,'' {\em Nature Communications}, vol.~14, Feb. 2023.

\bibitem{Pennartz2023}
C.~M.~A. Pennartz, M.~N. Oude~Lohuis, and U.~Olcese, ``How ‘visual’ is the visual cortex? the interactions between the visual cortex and other sensory, motivational and motor systems as enabling factors for visual perception,'' {\em Philosophical Transactions of the Royal Society B: Biological Sciences}, vol.~378, Aug. 2023.

\bibitem{Movshon2014}
J.~A. Movshon and E.~P. Simoncelli, ``Representation of naturalistic image structure in the primate visual cortex,'' {\em Cold Spring Harbor Symposia on Quantitative Biology}, vol.~79, p.~115–122, 2014.

\bibitem{MontesLourido2021}
P.~Montes-Lourido, M.~Kar, S.~V. David, and S.~Sadagopan, ``Neuronal selectivity to complex vocalization features emerges in the superficial layers of primary auditory cortex,'' {\em PLOS Biology}, vol.~19, p.~e3001299, June 2021.

\bibitem{critical-learning-periods}
A.~Achille, M.~Rovere, and S.~Soatto, ``Critical learning periods in deep neural networks,'' {\em arXiv preprint arXiv:1711.08856}, 2017.

\bibitem{Sotskov2022}
V.~P. Sotskov, N.~A. Pospelov, V.~V. Plusnin, and K.~V. Anokhin, ``Calcium imaging reveals fast tuning dynamics of hippocampal place cells and ca1 population activity during free exploration task in mice,'' {\em International Journal of Molecular Sciences}, vol.~23, p.~638, Jan. 2022.

\bibitem{log_dynamic}
G.~Buzsáki and K.~Mizuseki, ``The log-dynamic brain: how skewed distributions affect network operations,'' {\em Nature Reviews Neuroscience}, vol.~15, p.~264–278, Feb. 2014.

\bibitem{Taming-transformers}
P.~Esser, R.~Rombach, and B.~Ommer, ``Taming transformers for high-resolution image synthesis,'' in {\em Proceedings of the IEEE/CVF conference on computer vision and pattern recognition}, pp.~12873--12883, 2021.

\bibitem{Panzeri2015}
S.~Panzeri, J.~H. Macke, J.~Gross, and C.~Kayser, ``Neural population coding: combining insights from microscopic and mass signals,'' {\em Trends in Cognitive Sciences}, vol.~19, p.~162–172, Mar. 2015.

\bibitem{Gallego2017}
J.~A. Gallego, M.~G. Perich, L.~E. Miller, and S.~A. Solla, ``Neural manifolds for the control of movement,'' {\em Neuron}, vol.~94, p.~978–984, June 2017.

\bibitem{repE}
A.~Zou, L.~Phan, S.~Chen, J.~Campbell, P.~Guo, R.~Ren, A.~Pan, X.~Yin, M.~Mazeika, A.-K. Dombrowski, S.~Goel, N.~Li, M.~J. Byun, Z.~Wang, A.~Mallen, S.~Basart, S.~Koyejo, D.~Song, M.~Fredrikson, J.~Z. Kolter, and D.~Hendrycks, ``Representation engineering: A top-down approach to ai transparency,'' 2023.

\bibitem{templeton2024scaling}
A.~Templeton, T.~Conerly, J.~Marcus, J.~Lindsey, T.~Bricken, B.~Chen, A.~Pearce, C.~Citro, E.~Ameisen, A.~Jones, H.~Cunningham, N.~L. Turner, C.~McDougall, M.~MacDiarmid, C.~D. Freeman, T.~R. Sumers, E.~Rees, J.~Batson, A.~Jermyn, S.~Carter, C.~Olah, and T.~Henighan, ``Scaling monosemanticity: Extracting interpretable features from claude 3 sonnet,'' {\em Transformer Circuits Thread}, 2024.

\bibitem{AlejandreGarca2022}
T.~Alejandre-García, S.~Kim, J.~Pérez-Ortega, and R.~Yuste, ``Intrinsic excitability mechanisms of neuronal ensemble formation,'' {\em eLife}, vol.~11, May 2022.

\bibitem{Bau-2020-role-of-individual-neurons}
D.~Bau, J.-Y. Zhu, H.~Strobelt, A.~Lapedriza, B.~Zhou, and A.~Torralba, ``Understanding the role of individual units in a deep neural network,'' {\em Proceedings of the National Academy of Sciences}, vol.~117, no.~48, pp.~30071--30078, 2020.

\bibitem{matveev2021overview}
S.~A. Matveev, I.~V. Oseledets, E.~S. Ponomarev, and A.~V. Chertkov, ``Overview of visualization methods for artificial neural networks,'' {\em Computational Mathematics and Mathematical Physics}, vol.~61, no.~5, pp.~887--899, 2021.

\bibitem{selvaraju2017grad}
R.~R. Selvaraju, M.~Cogswell, A.~Das, R.~Vedantam, D.~Parikh, and D.~Batra, ``Grad-cam: Visual explanations from deep networks via gradient-based localization,'' in {\em Proceedings of the IEEE international conference on computer vision}, pp.~618--626, 2017.

\bibitem{sundararajan2017axiomatic}
M.~Sundararajan, A.~Taly, and Q.~Yan, ``Axiomatic attribution for deep networks,'' in {\em International conference on machine learning}, pp.~3319--3328, PMLR, 2017.

\bibitem{goh2021multimodal}
G.~Goh, N.~Cammarata, C.~Voss, S.~Carter, M.~Petrov, L.~Schubert, A.~Radford, and C.~Olah, ``Multimodal neurons in artificial neural networks,'' {\em Distill}, vol.~6, no.~3, p.~e30, 2021.

\bibitem{Arakawa2008}
H.~Arakawa, D.~C. Blanchard, K.~Arakawa, C.~Dunlap, and R.~J. Blanchard, ``Scent marking behavior as an odorant communication in mice,'' {\em Neuroscience; Biobehavioral Reviews}, vol.~32, p.~1236–1248, Sept. 2008.

\bibitem{HSE-Supercomputer-counter-paper}
P.~Kostenetskiy, R.~Chulkevich, and V.~Kozyrev, ``Hpc resources of the higher school of economics,'' in {\em Journal of Physics: Conference Series}, vol.~1740, p.~012050, IOP Publishing, 2021.

\end{thebibliography}

\appendix

\section{Gallery of notable MEIs}
\label{appendix:MEI-gallery}

In this section we present some notable MEIs for neurons of different (spiking) layers and attempt to interpret their specializations. 

\begin{figure}[H]
\centering
\includegraphics[width=0.5\textwidth]{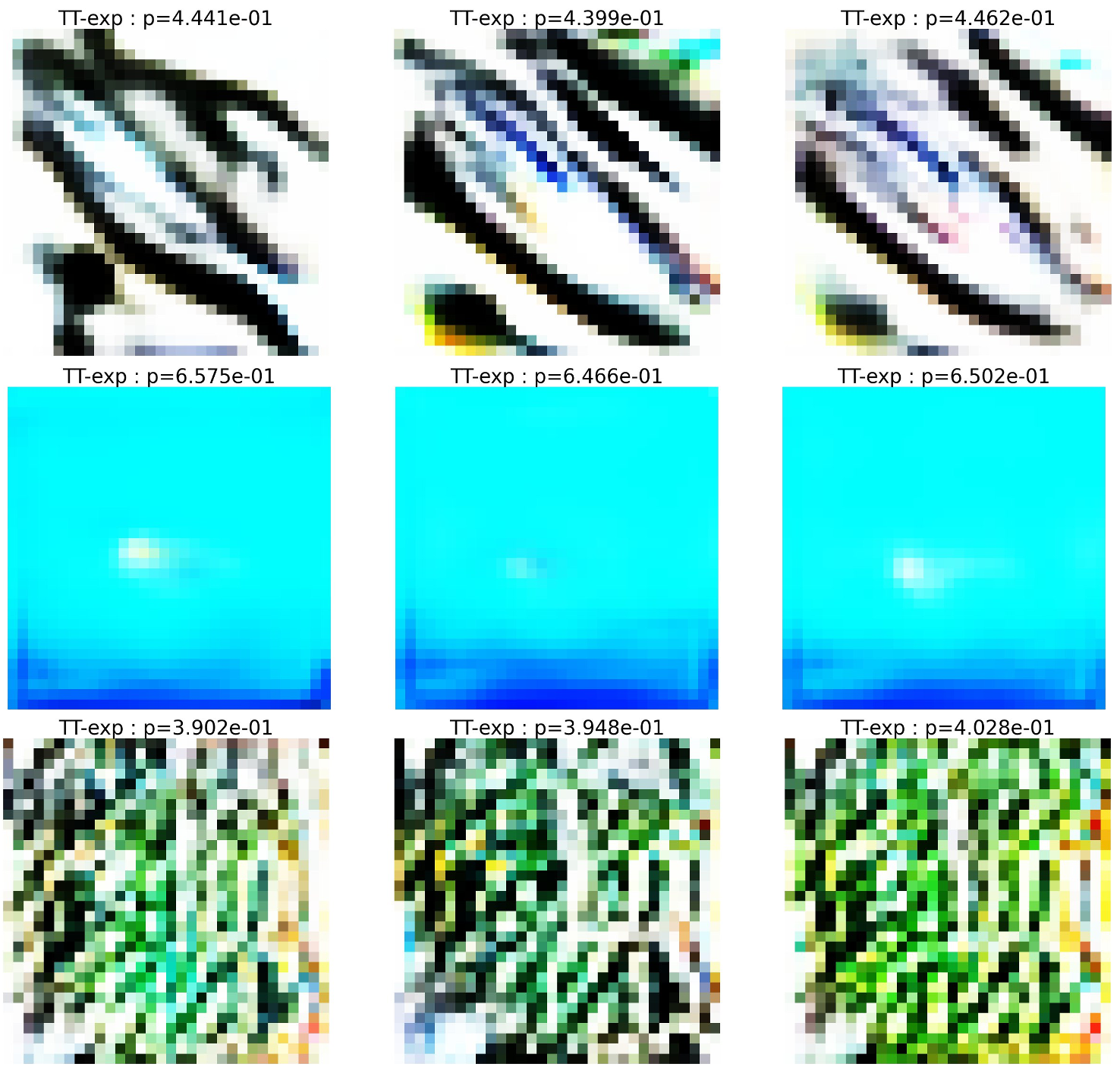}
\caption{MEIs of first spiking (sn1) layer neurons. \textbf{Each row} is a certain neuron, several activation optima.}
\label{fig:MEI-gallery-0}
\end{figure}

Fig.\thinspace\ref{fig:MEI-gallery-0} shows MEIs of neurons of the first spiking layer (sn1). Despite the spiking nature of neurons, these clearly specialize on visual primitives – patterns/textures, uniform monochrome ``blobs'' – just like non-spiking neurons in feedforward ANNs \cite{Bengio-transferable-features,AlexNet}. See Figs.\thinspace\ref{fig:argmax-gen-comp-color},\thinspace\ref{fig:argmax-gen-comp-stripes} in Appendix \ref{appendix:VAE} for more example of first layer neurons' MEIs. MEIs get more interesting with layer depth.

\begin{figure}[H]
\centering
\includegraphics[width=1.0\textwidth]{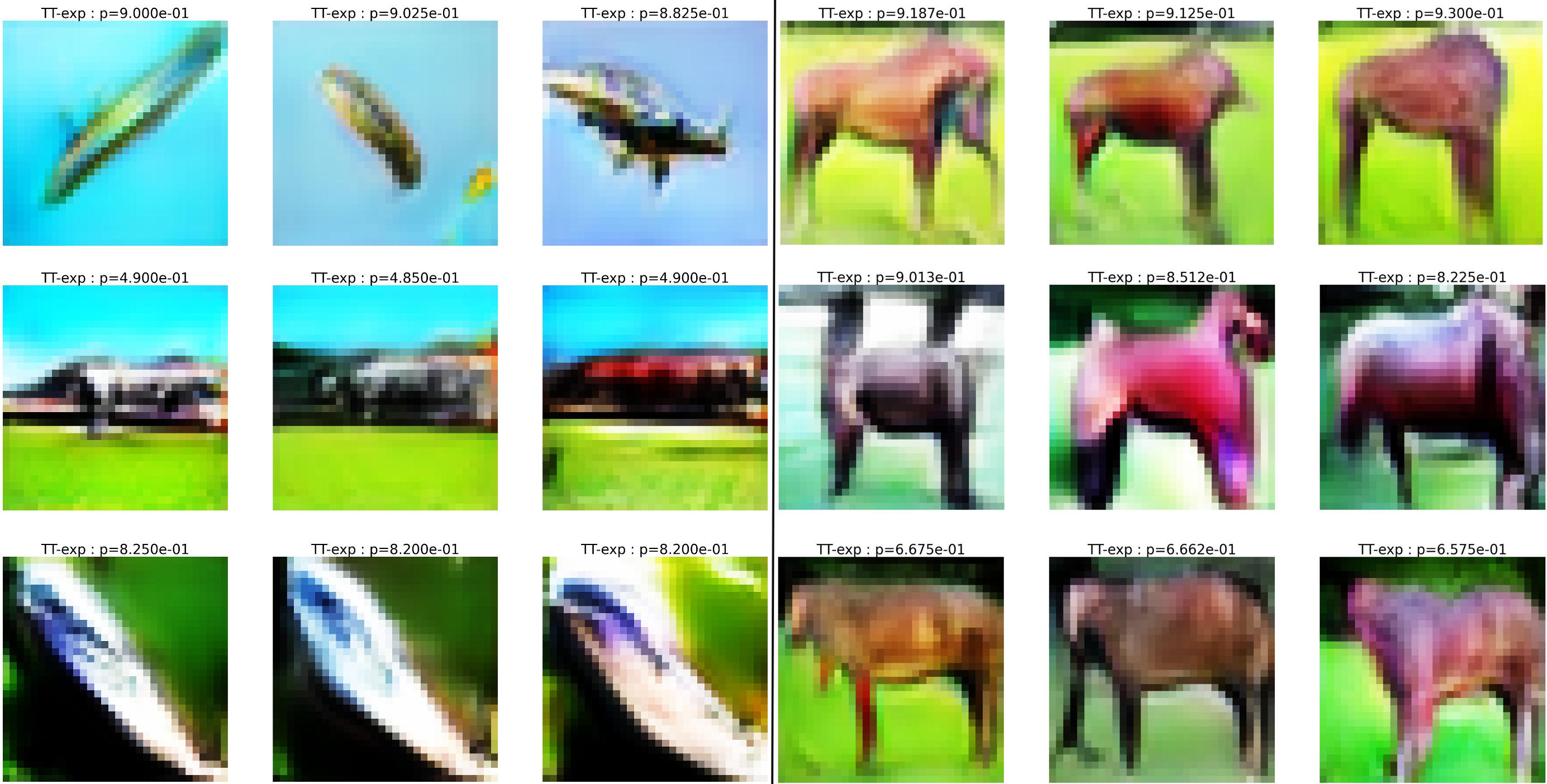}
\caption{MEIs of intermediary (layer2.1.sn1) layer neurons. \textbf{Each rows of 3 images:} one neuron, several activation optima. \textbf{Left 3x3 block:} 3 different neurons with different specializations. \textbf{Right 3x3 block:} 3 different neurons with similar specializations.}
\label{fig:MEI-gallery-1}
\end{figure}

Fig.\thinspace\ref{fig:MEI-gallery-1} shows MEIs of different neurons of an moderately deep layer. In the left 3x3 block, first row neuron specializes on elongated shapes on blue background: MEIs are close to typical images of classes 8 – ship and 0 – airplane. Second row neuron specializes on green bottom and blue top background (grass and skies) with objects in between (automobiles / trucks / airplanes on runway strips). Third row neuron is possibly the most interesting, as it seems to specialize on images sharing features of birds and ships (green freshwater) – we call it the \textbf{bird-ship neuron}. Also see Section.\thinspace\ref{subsec:emergence-and-dynamics} and Fig.\thinspace\ref{fig:labile-neurons-distr-and-bird-ship}.

In the right 3x3 block, we demonstrate 3 \textbf{different neurons with very similar specializations}. Clearly the MEIs share the features of dataset classes 7 – horse, 4 – deer (possibly also 5 – dog), but were often found to be almost mirror-symmetric, so it seems these neurons specialize on such ``\textbf{brown tetrapods}'' on green (grass) background without much knowledge of head vs tail placement. Overall it is expected that moderately deep neurons specialize on such simple image types – more complicated than visual primitives, but less complicated than detailed images clearly distinguishable as dataset classes. What is notable however is that such smooth but not yet very class-specific MEIs are stably obtained with a gradient-free optimization procedure, not by simply back-propagating as in feedforward ANNs \cite{Olah-feature-visualization}.

\begin{figure}[H]
\centering
\includegraphics[width=0.5\textwidth]{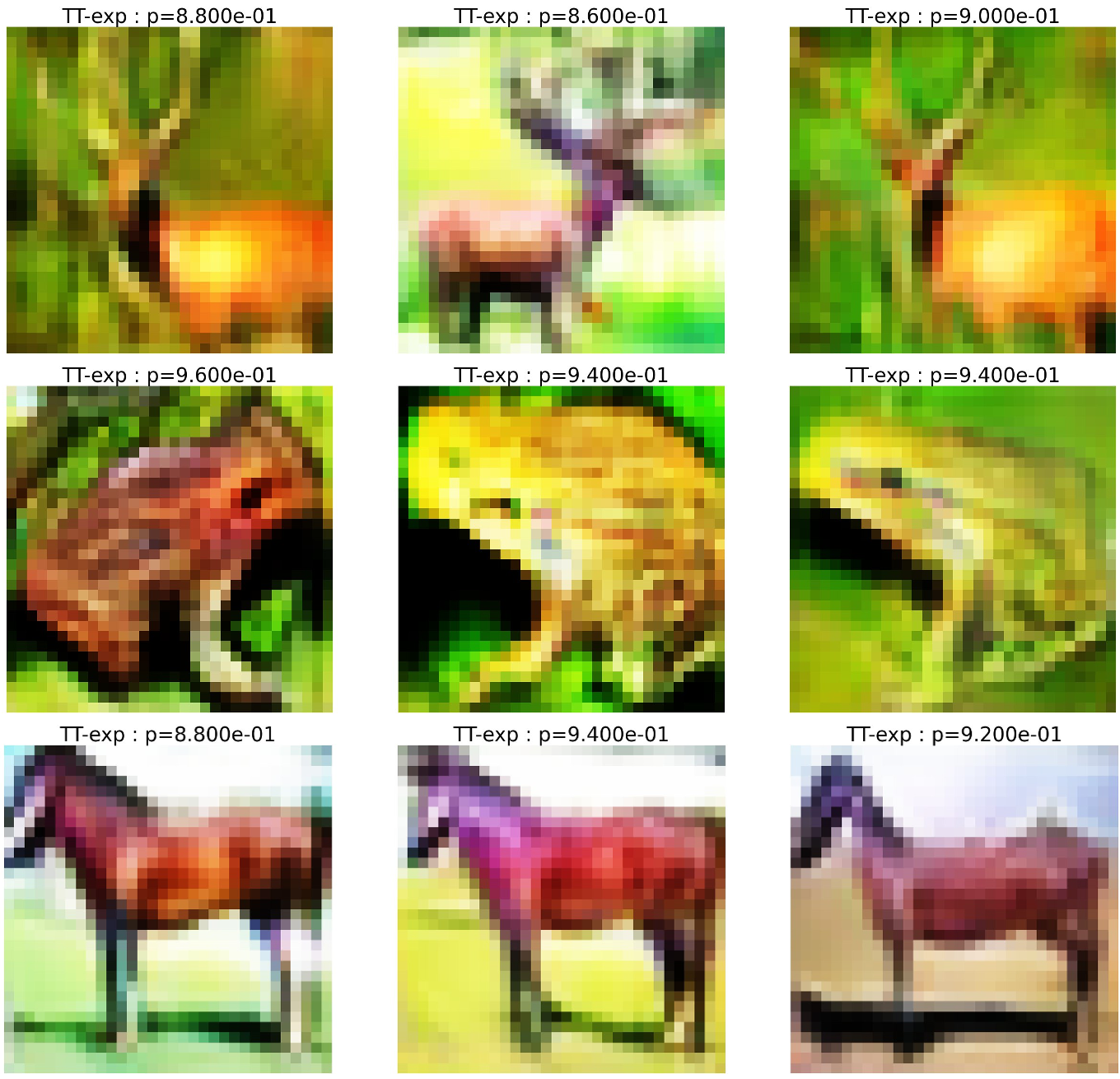}
\caption{MEIs of last spiking layer (layer4.1.sn1). \textbf{Each row} is a certain neuron, several activation optima.}
\label{fig:MEI-gallery-2}
\end{figure}

Figure\thinspace\ref{fig:MEI-gallery-2} shows MEIs of the last (deepest) spiking layer (layer4.1.sn1). Its neurons clearly specialize on dataset-like images, with classes 4 – deer, 6 – frog, 7 – horse – clearly distinguishable.

\section{Generative models comparison}\label{appendix:VAE}

\subsection{CIFAR-10 dataset}

\begin{figure}[H]
\centering
\includegraphics[width=0.75\textwidth]{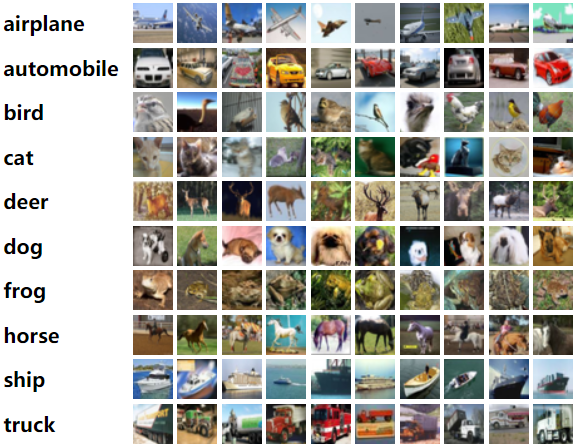}
\caption{10 classes of images in CIFAR10 dataset \cite{CIFAR10-dataset-reference}}
\label{fig:CIFAR10-classes}
\end{figure}

The models were trained on CIFAR-10  \cite{CIFAR10-dataset-reference} – a dataset of 60.000 color images of size 32x32 pixels, split into 10 classes (6.000 images per class): 0) airplane, 1) automobile, 2) bird, 3) cat, 4) deer, 5) dog, 6) frog, 7) horse, 8) ship, 9) truck. Figure\thinspace\ref{fig:CIFAR10-classes} provides examples of dataset images.

\subsection{Overview of VAEs}

Apart from SN-GANs, we have also tested VQ-VAEs for image generation. A Vector-Quantized Variational AutoEncoder (VQ-VAEs) \cite{VQ-VAE-Van-Den-Oord} is a generative model of the VAE family \cite{VAE-vanilla-Kingma-Welling}. VAEs were chosen for our task mainly due being quick to generate data – only one forward pass is needed to obtain a sample; but also due to simplicity of architecture – encoder-decoder, where the endoder gives the latent coordinates of a sample.

VAEs take a sample of data, $x$ (e.g. a batch of images) and transform it, layer by layer of the encoder (if data is images, some convolutional layers are applied first), to the latent space. On that space, a family of parameterized probability distributions (posteriors) $p_\theta (z|x)$ is defined (one can say that the encoder just outputs a vector of parameters $\theta$ of this distribution, thus specifying a representative of this family). One then generates a sample $z\sim p_\theta(z|x)$ from that distribution – this is a non-deterministic action and so at first it seems non-differentiable for one to design a backward pass (error backpropagation), but a clever idea of the reparameterization trick \cite{VAE-vanilla-Kingma-Welling} overcomes this problem (by separating parameterization of a distribution from sampling from it). This ``latent code'' $z$ that describes the input sample $x$ is then passed to the decoder (if the VAE has to generate images – last layers of the decoder will be deconvolutional, etc.) to produce an output – a new sample of data. There is also what's called a prior distribution $p(z)$ of latent codes, which, for the case of latent space being $\mathbb{R}^d$, is uninformatively chosen to be standard normal (uniform distribution can't be supported on $\mathbb{R}^d$). There exists a problem of ``posterior collapse'' – when the posterior distribution gets too close to the uninformative prior on some latent codes. Various modifications of VAEs aim to fix this problem, with VQ-VAE doing so quite successfully under certain conditions. 

Vector-Quantized VAEs \cite{VQ-VAE-Van-Den-Oord} take this idea to a ``discrete'' setting – the family of posterior distributions $p_\theta(z|x)$ is now discrete rather than continuous (multidimensional normal in original VAEs), supported on a fixed-size \textbf{dictionary of latent codes} (codewords). The motivation behind this is that, for any finite sample of natural images, there will only be a discrete set of classes in it – dogs, cats, cars, etc. So VQ-VAEs tend to produce far sharper images than their original VAE counterparts (which tend to blur the image, with the uncertainty of the posterior distribution translating into gaussian-like, albeit nonlinearly transformed, blur on the resulting images), which at first seemed better for our needs. The encoder of VQ-VAE thus learns to output these discrete posteriors (the latent codes and their posterior probabilities) so that samples from them (output of the decoder) were alike the training data. Since the prior $p(z)$ is also supported on this discrete codebase (which can support a uniform distribution), choosing it to be uniform helps prevent posterior collapse.  

In our experiments, we've tested VQ-VAEs with \textbf{latent dimension 64} with \textbf{dictionary size of 512} – while CIFAR-10 only has 10 classes, codes are assigned not to singular images, but to their batches (samples). Both encoder and decoder were built with ResidualStacks \cite{ResNet-paper} of 3-4 convolutional layers.

\subsection{Comparison of GAN- and VAE- based generators}

For MEIs created for each neuron of all spiking layers from the spiking ResNet18, analysis has shown that the maximal activations of neurons on images generated by VQ-VAE were smaller than on images genereted by SN-GAN. The magnitude of the gap depended on the layer and ranged from 50\% in early layers to 5\% in deep layers.

In general, despite neuronal activations were significantly above chance, the MEIs generated by VAE often contain no visible structure and do not provide any insight about neuronal specializations.

The exception is the early layers, in which the MEIs obtained using both generators are quite similar. This allows one expect that discrete optimization methods actually find \textbf{generator-agnostic maxima} in the global image space, and difficulties in obtaining interpretable images can be solved by further improving the structure of generators' latent space.
%(Fig. \ref{fig:argmax-gen-comp-color})
\begin{figure}[H]
\centering
\includegraphics[width=0.5\textwidth]{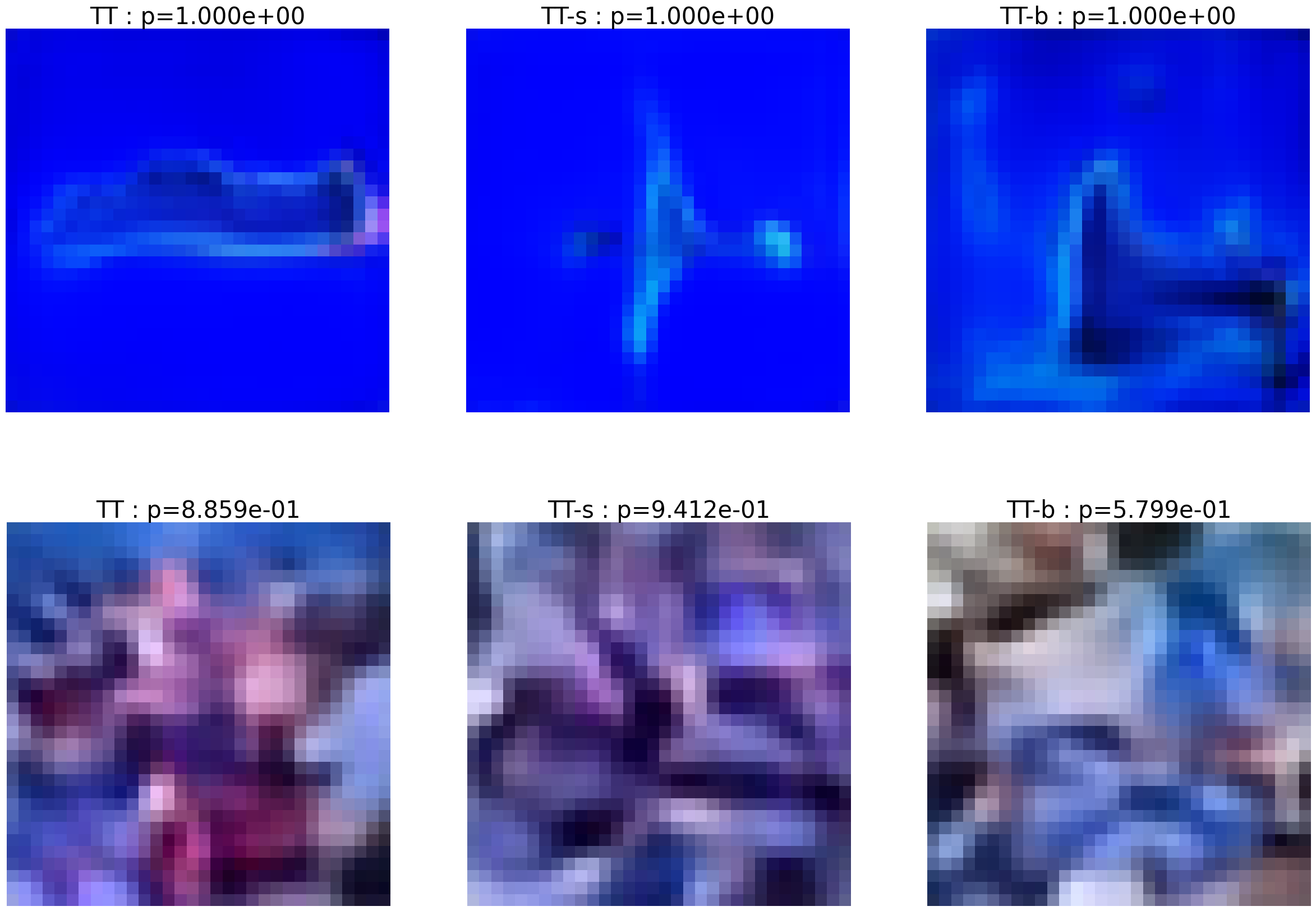}
\caption{MEIs for unit 16 of the first spiking layer from spiking ResNet18. \textbf{Top:} images generated with SN-GAN. \textbf{Bottom:} images generated with VQ-VAE. Columns correspond to different TT-based optimization methods.}
\label{fig:argmax-gen-comp-color}
\end{figure}

\begin{figure}[H]
\centering
\includegraphics[width=0.5\textwidth]{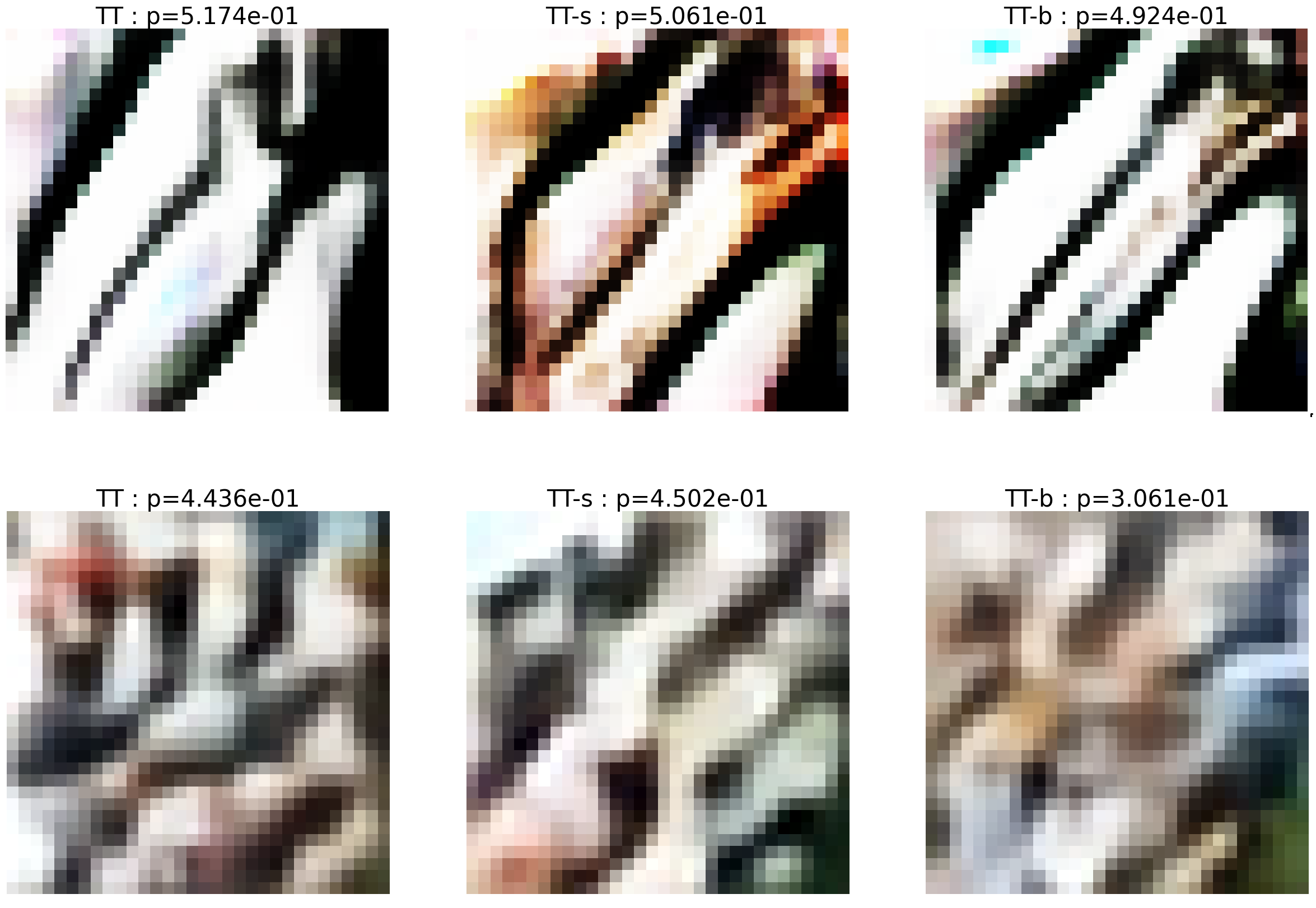}
\caption{MEIs for unit 35 of the first spiking layer from spiking ResNet18. \textbf{Top:} images generated with SN-GAN. \textbf{Bottom:} images generated with VQ-VAE. Columns correspond to different TT-based optimization methods.}
\label{fig:argmax-gen-comp-stripes}
\end{figure}

Thus, VQ-VAE-generated images are closest to richer SN-GAN-generated ones where there are uniform colors or simple geometric patterns in the image (see Figs\thinspace\ref{fig:argmax-gen-comp-color}, ~\ref{fig:argmax-gen-comp-stripes}). We suppose that this is a sign of excessive information compression in the latent space of VQ-VAE generator, which leads to the loss of significant image details.

Our hypothesis as for why VQ-VAE is performing poorer in our setup is: as described in the above sections, VQ-VAEs learn discretized, rather than continuous posterior distributions of data samples – supported on a finite set (dictionary) of (learned) vectors in the latent space, the posterior distribution then being multinomial on that discrete pointset. For that reason, VQ-VAEs seem to be not so good at interpolating between probable datapoints (images) – designed to mitigate posterior collapse and gaussian-like blur of generated images in original VAEs, it seems that VQ-VAEs loose in the ability to generate interesting out-of-distribution samples. Of course, some interpolation is possible, and it seems we didn't find good balance between discretization parameters of VQ-VAE and TT – that is matter of future investigation. 

SN-GANs performed far better in our experiment, providing seemingly smoother (due to benefits of Spectral Normalization) coordinates on the latent space of images. However, the above results on neuron specialization distributions seem to be universal – no matter what generative model was used.

\end{document}